\documentclass[table,dvipsnames]{fairmeta}
\usepackage{amsmath}
\usepackage{enumerate} 
\usepackage{algorithm}
\usepackage{algpseudocode}
\usepackage{amsfonts}
\usepackage{amsthm}
\usepackage{newtxtt}
\usepackage{algorithm}
\usepackage{bm}
\usepackage{diagbox} 
\usepackage{colortbl}
\usepackage{nicematrix}
\usepackage{subcaption}
\usepackage{cellspace}
\usepackage{tcolorbox}
\usepackage[framemethod=tikz]{mdframed}
\usepackage{amssymb}
\usepackage{xspace}
\usepackage{wrapfig}
\usepackage{adjustbox}
\usepackage{tabularx}
\usepackage{booktabs}
\usepackage{mathtools}
\usepackage{tikz}
\usetikzlibrary{arrows.meta, bending} %
\usetikzlibrary{decorations.text}

\definecolor{pink}{RGB}{255, 204, 204}
\definecolor{bluel}{RGB}{230, 230, 255}
\definecolor{light}{RGB}{125, 125, 125}
\definecolor{pastelblue}{HTML}{cee4f6}
\definecolor{pastelgreen}{RGB}{204, 255, 204}
\definecolor{pastelpink}{HTML}{edc8e3}

\definecolor{darkerpink}{RGB}{220, 110, 110}
\definecolor{pastelgray}{RGB}{237, 237, 237}
\definecolor{faintred}{HTML}{FFDDDD}
\definecolor{lightred}{HTML}{FFBBBB}
\definecolor{darkred}{HTML}{FF9999}

\usepackage{silence}

\definecolor{mygray}{gray}{0.95}

\newcommand{\er}[1]{\color{Gray} {\tiny +0.0\%}}

\theoremstyle{plain}

\theoremstyle{definition}

\theoremstyle{remark}

\makeatletter
\newtheorem*{rep@theorem}{\rep@title}
\newcommand{\newreptheorem}[2]{%
\newenvironment{rep#1}[1]{%
 \def\rep@title{\textbf{#2} \ref{##1}}%
 \begin{rep@theorem}}%
 {\end{rep@theorem}}}
\makeatother

\newreptheorem{theorem}{Theorem}
\newreptheorem{proposition}{Proposition}
\newreptheorem{lemma}{Lemma}
\newreptheorem{corollary}{Corollary}

\def\eqref#1{equation~\ref{#1}}

\def\1{\bm{1}}

\crefformat{equation}{#2equation~#1#3}
\Crefformat{equation}{#2Equation~#1#3}

\DeclareMathAlphabet{\mathsfit}{\encodingdefault}{\sfdefault}{m}{sl}
\SetMathAlphabet{\mathsfit}{bold}{\encodingdefault}{\sfdefault}{bx}{n}

\newcolumntype{C}[1]{>{\Centering}m{#1}}

\newcolumntype{Z}[1]{>{\Left}m{#1}}

\algdef{SE}[REPEATN]{RepeatN}{End}[1]{\algorithmicrepeat\ #1 \textbf{times}}{\algorithmicend \textbf{ repeat}}

\usepackage[frozencache,cachedir=.]{minted}
\definecolor{LightGray}{gray}{0.97}
\setminted{
    frame=lines,
    framesep=2mm,
    baselinestretch=1.2,
    bgcolor=LightGray,
    fontsize=\footnotesize,
    breaklines,
    linenos
}

\newmdenv[backgroundcolor=metabg, roundcorner=5pt, skipabove=7pt, linewidth=0pt, innertopmargin=4pt]{myframe}

\renewcommand{\eqref}[1]{\labelcref{#1}}

\usepackage{bbding}
\usepackage{amssymb}
\usepackage{booktabs, makecell, tabularx}
\usepackage{tabularray}
\usepackage{tikz}
\usepackage{algorithm}
\usepackage{dsfont}
\usepackage{diagbox}

\title{StreamDiffusionV2: A Streaming System for Dynamic and Interactive Video Generation}

\author[]{Tianrui Feng$^{1}$}
\author[]{Zhi Li$^{2}$}
\author[]{Shuo Yang$^{2}$}
\author[]{Haocheng Xi$^{2}$}
\author[]{Muyang Li$^{3}$}
\author[]{Xiuyu Li$^{2}$}
\author[]{Lvmin Zhang$^{4}$}
\author[]{Keting Yang$^{5}$}
\author[]{Kelly Peng$^{6}$}
\author[]{Song Han$^{7}$}
\author[]{Maneesh Agrawala$^{4}$}
\author[]{Kurt Keutzer$^{2}$}
\author[]{Akio Kodaira$^{8}$}
\author[]{Chenfeng Xu$^{\dagger, 1}$}

\affiliation[]{$^{1}$UT Austin}
\affiliation[]{$^{2}$UC Berkeley}
\affiliation[]{$^{3}$Nunchaku AI}
\affiliation[]{$^{4}$Stanford University}
\affiliation[]{$^{5}$Independent Researcher}
\affiliation[]{$^{6}$First Intelligence}
\affiliation[]{$^{7}$MIT}
\affiliation[]{$^{8}$Shizuku AI}

\abstract{Generative models are reshaping the live-streaming industry by redefining how content is created, styled, and delivered. Previous image-based streaming diffusion models have powered efficient and creative live streaming products but has hit limits on temporal consistency due to the foundation of image-based designs.
Recent advances in video diffusion have markedly improved temporal consistency and sampling efficiency for offline generation. However, offline generation systems primarily optimize throughput by batching large workloads. In contrast, live online streaming operates under strict service-level objectives (SLOs): time-to-first-frame must be minimal, and every frame must meet a per-frame deadline with low jitter. Besides, scalable multi-GPU serving for real-time streams remains largely unresolved so far. To address this, we present \textbf{StreamDiffusionV2}, a \emph{training-free} pipeline for interactive live streaming with video diffusion models. StreamDiffusionV2 integrates an SLO-aware batching scheduler and a block scheduler, together with a sink-token–guided rolling KV cache, a motion-aware noise controller, and other system-level optimizations. Moreover, we introduce a scalable pipeline orchestration that parallelizes the diffusion process across denoising steps and network layers, achieving near-linear FPS scaling without violating latency guarantees. The system scales seamlessly across heterogeneous GPU environments and supports flexible denoising steps (e.g., 1–4), enabling both ultra-low-latency and higher-quality modes. Without TensorRT or quantization, StreamDiffusionV2 renders the first frame within 0.5s and attains 58.28 FPS with a 14B-parameter model and 64.52 FPS with a 1.3B-parameter model on four H100 GPUs. Even when increasing denoising steps to improve quality, it sustains 31.62 FPS (14B) and 61.57 FPS (1.3B), making state-of-the-art generative live streaming practical and accessible—from individual creators to enterprise-scale platforms.

\vskip 0.05in
\centering{Project page: \url{https://streamdiffusionv2.github.io/}} \\
\vskip 0.05in
\textsuperscript{$\dagger$}Project lead, corresponding to {xuchenfeng@utexas.edu}
\vskip 0.05in
}

\begin{document}

\maketitle

\section{Introduction}\label{section:intro}

\begin{figure*}[ht]
\centering
\includegraphics[width=\linewidth]{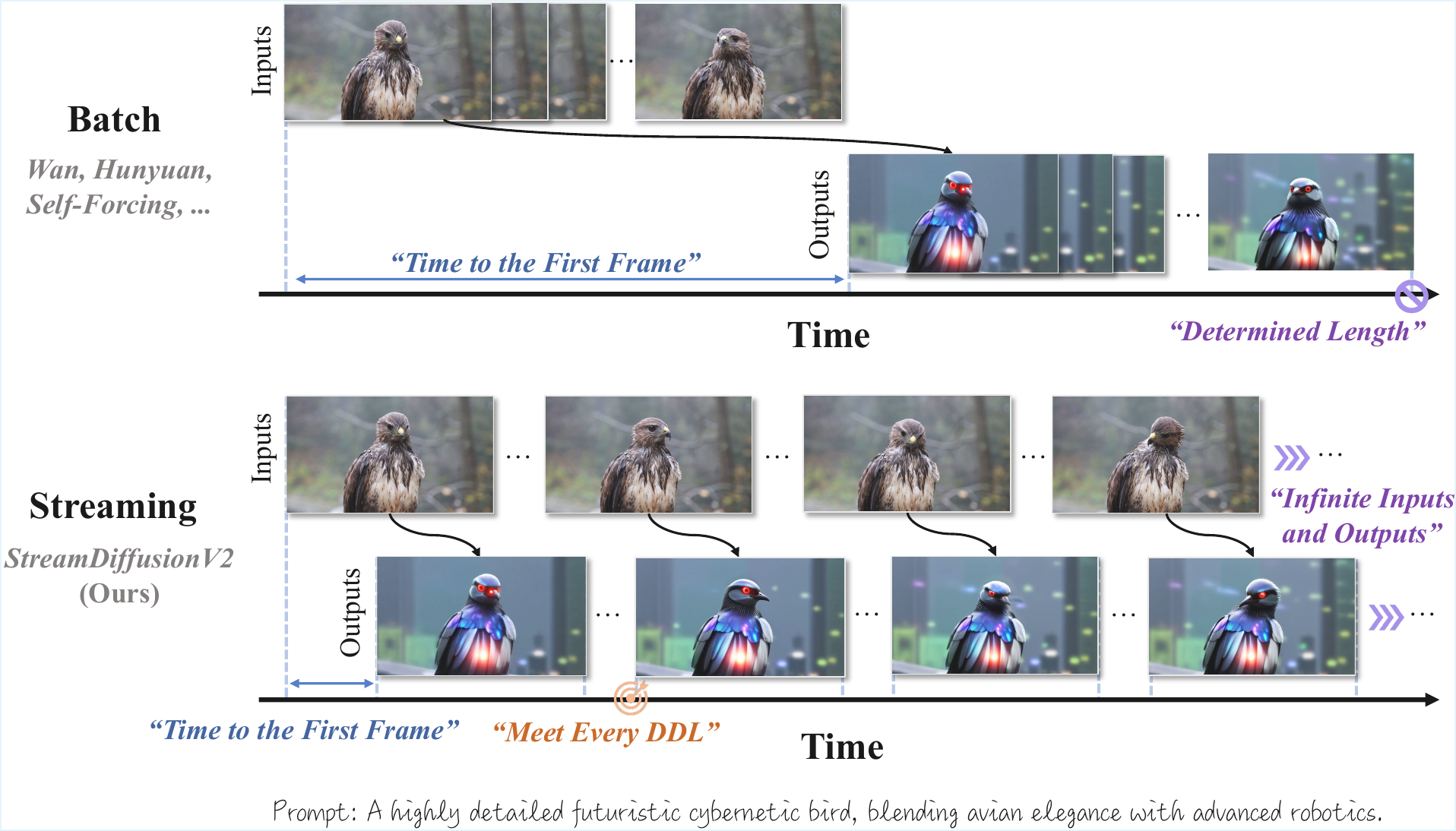}
\caption{\textbf{Comparison between Batch and Streaming video generation.} Different from generating a large batch of video, live-streaming video generation targets at cutting down the ``time to the first frame'' and producing continuous output with low latency.}
\label{fig: stream}
\vspace{-10pt}
\end{figure*}

Recent advances in diffusion models have transformed the way live video content is created and rendered. Modern live AI streaming pipelines, such as Daydream\footnote{\url{https://blog.livepeer.org/introducing-daydream/}} and TouchDesigner\footnote{\url{https://derivative.ca/tags/streamdiffusion}}, are largely powered by image-diffusion–based methods~\citep{kodaira2023streamdiffusion,liang2024looking} due to their flexible style control and responsiveness and ease of integration with real-time applications. With simple text prompts, creators can restyle scenes, replace backgrounds, add effects, and even animate virtual hosts that respond to live chat. These pipelines can also repurpose the same input footage into derivative formats such as short clips and thumbnails, dramatically reducing post-production effort. By combining low cost, high speed, and interactive visual fidelity, image-diffusion streaming has redefined traditional live-streaming workflows and enabled new creative use cases in gaming, social media, and live entertainment.

However, their image-centric design exposes a fundamental weakness: poor temporal consistency. Frame-by-frame generation introduces flicker and drift, causing noticeable instability across time. 
In practice, even commercial-grade systems still struggle to maintain coherent motion and appearance in continuous streams.

In parallel, video diffusion models \citep{huang2025self,yin2025slow,wan2025wan} have achieved far stronger temporal consistency through explicit modeling of temporal dependencies. Recent efforts have pushed these models towards fast, even ``real-time'' generation~\citep{xi2025sparse,yang2025sparse,yang2025longlive}, making them promising candidates for streaming. A natural question arises: can these models replace image-based diffusion in real-time AI video pipelines?

Despite their efficiency improvements, current video diffusion systems remain fundamentally ill-suited for live streaming. Our systematic analysis identifies four key challenges:
\begin{itemize}
    \item \textbf{Unmet real-time SLOs.}
    State-of-the-art video diffusion models (e.g., WAN~\citep{wan2025wan}) and their efficient variants~\citep{huang2025self,yin2025slow} are optimized for offline generation throughput by processing large, fixed inputs of shape \(1 \times T \times H \times W\) per forward pass, where \(T\) often ranges from 81 to hundreds of frames. Such large chunks violate real-time service-level objectives (SLOs)~\citep{sripanidkulchai2004feasibility, zhang2005coolstreaming, huang2008challenges}, which requires minimal time-to-first-frame (TTFF) and strict per-frame deadlines (see Fig.\,\ref{fig: stream}). The fixed input size also prevents adaptive scheduling under varying hardware load, making SLO compliance difficult across heterogeneous GPU environments.

    \item \textbf{Drift over unbounded horizons.} Existing video diffusion pipelines generate fixed-length clips using static configurations, including KV caches, sink tokens, and RoPE schedules, calibrated for bounded temporal contexts. These assumptions break down in continuous live streams, where temporal context, content statistics, and user prompts evolve dynamically. Over long horizons (e.g., hour-long session), these static components accumulate misalignment, leading to visual drift and degraded temporal coherence.

    \item \textbf{Motion tearing under high-speed dynamics.} Current video diffusion models are predominantly trained on slow-motion datasets~\citep{bain2021frozen, chen2024panda, yang2024vript}, leaving them poorly adapted to fast motion. Their inference pipelines often use large chunk sizes with rule-based noise schedules to suppress inter-frame variation, which over-smooths motion and erodes fine details. As a result, live streams that feature rapid camera movement or action sequences suffer from blur, ghosting, and motion tearing. 

    \item \textbf{Poor GPU scaling.}
    As AI-driven live streaming scales from individual creators to enterprise-grade deployments, multi-GPU scaling becomes critical. However, existing offline diffusion parallelization strategies do not translate to real-time workloads. Sequence parallelism suffers from unpredictable communication overhead, while naive pipeline parallelism yields minimal improvement in FPS. Because per-frame latency constraints dominate real-time workloads, achieving scalable, low-jitter performance remains an open system challenge.

\end{itemize}
These challenges call for a new inference system designed for real-time constraints rather than offline throughput.

We introduce \textbf{StreamDiffusionV2}, a training-free streaming system that adapts video diffusion models for interactive, low-latency generation. Our objective is stringent: \textit{to satisfy live-streaming SLOs — low time-to-first-frame and strict per-frame deadlines (DDL) — while preserving temporal consistency and visual fidelity over long, dynamic sequences, and more importantly, to provide a scalable solution that serves users at different levels of compute capacity.} 

StreamDiffusionV2 synergizes several techniques to achieve both efficiency and visual quality objectives.
\textbf{Efficiency objectives:}
\textbf{(1)} To satisfy live-streaming SLOs on a single GPU, instead of using a fixed input of $1 \times T \times H \times W$, we employ an SLO-aware \emph{batching scheduler} that reformulates inputs as $B \times T' \times H \times W$. We intentionally keep $T'$ small (e.g., only a few frames) to limit per-step latency and meet per-frame deadlines (DDL), while adapting the \emph{stream batch} size $B$~\citep{kodaira2023streamdiffusion} to instantaneous hardware load to maximize utilization.
\textbf{(2)} To deliver scalable performance across multiple GPUs, we introduce a dynamic scheduler that orchestrates the pipeline across both denoising steps and network stages. Naive pipeline parallelism alone does not yield linear FPS scaling with additional GPUs. We pair it with the SLO-aware batching scheduler above to keep all devices well utilized under SLO-aware workloads, thereby improving aggregate FPS while maintaining latency guarantees.
\textbf{Vision quality objectives:}
\textbf{(1)} To support unbounded live-streaming use cases, we continuously update the sink tokens to reflect the current prompt semantics and recent visual context, and we refresh anchor caches when topics or motion regimes change. We also reset RoPE offsets at chunk boundaries to avoid positional misalignment over long horizons. Together, these controls preserve appearance and motion semantics while retaining the latency benefits.
\textbf{(2)} To handle high-speed motion, we introduce a motion-aware noise scheduler that estimates motion magnitude (e.g., via lightweight optical-flow proxies) and adapts the noise level and timestep schedule on a per-chunk basis. Specifically, fast motion receives more conservative denoising to suppress tearing and ghosting, while slow motion benefits from more aggressive refinement to recover fine details. This design substantially improves sharpness and temporal stability for high-speed content in live streams.

StreamDiffusionV2 integrates the aforementioned multiple system-level techniques into a cohesive, training-free pipeline that transforms efficient video diffusion models into real-time, stream-live applications. These components enable high FPS, high visual fidelity, and strong temporal consistency for AI-driven live streaming. Without relying on TensorRT or quantization, StreamDiffusionV2 is able to achieve time-to-first-frame within 0.5 seconds and is the first system to achieve 58.28 FPS with a 14B-parameter model and 64.52 FPS with a 1.3B-parameter model, both running on four H100 GPUs. Even when increasing the number of denoising steps to enhance generation quality, the system maintains 31.62 FPS (14B) and 61.57 FPS (1.3B). Beyond raw speed, StreamDiffusionV2 provides a tunable trade-off across resolution, denoising steps, and GPU scale, allowing users to balance quality, throughput, and resource constraints. This flexibility supports a wide range of deployment scenarios, from individual creators using a single GPU to enterprise-scale platforms operating large GPU clusters. StreamDiffusionV2 establishes a practical foundation for next-generation live generative media systems. We will release our implementation to promote open research and foster innovation in real-time interactive video generation.

\section{Related Works}
\paragraph{Efficient Offline Video Generation.}
Diffusion-based video generation achieves impressive visual fidelity, but the inference latency of Video DiTs remains a key bottleneck. Prior work tackles this from two complementary angles: training-based and training-free methods. Training-based approaches—such as distillation~\citep{salimans2022progressive, kim2023consistency, meng2023distillation, yin2024one, yin2024improved, lu2024simplifying} and linear attention~\citep{gao2024matten, chen2024deep, dalal2025one, xie2024sana, chen2025sana, po2025long}—optimize the model to shorten the diffusion process or reparameterize attention for faster inference. In contrast, training-free methods—including cache reuse~\citep{selvaraju2024fora, chen2024delta, wimbauer2024cache, liu2024timestep, zhou2025less} and sparse attention~\citep{xi2025sparse, yang2025sparse, zhang2025sageattention, zhang2024sageattention2, zhang2025spargeattn}—improve runtime by reusing latent features or sparsifying attention computation. Together, these techniques substantially reduce Video DiT latency in offline scenarios. However, despite strong offline speedups, they do not transfer directly to streaming, which requires unbounded (infinite) queues and strict low-latency, online generation.

\paragraph{Streaming Video Generation.}
Autoregressive
video generation~\citep{chen2025skyreels, teng2025magi, kodaira2025streamdit, yin2025slow, huang2025self, yang2025longlive} is well aligned with our streaming setting: the forward-only, continuous process produces frames sequentially with significantly faster speed compared to offline generation. 
Among these approaches, CausVid~\citep{yin2025slow} distills a student-initialized bidirectional DiT into a causal DiT, while Self-Forcing~\citep{huang2025self} employs self-rollout training to curb error accumulation. A series of works \citep{yang2025longlive,he2025matrix,kodaira2025streamdit,valevskidiffusion} then extend the idea of a few-step autoregressive paradigm to support infinite-length synthesis for fast streaming generation. Despite substantial progress toward live-streaming, existing methods still fall short of streaming SLOs. More importantly, although fast, both CausVid and Self-Forcing struggle with high-speed motion: their training biases toward motion smoothing can introduce temporal over-smoothing and visible tearing during streaming. To address these gaps, StreamDiffusionV2 introduces a serving pipeline that adapts state-of-the-art efficient video-diffusion models to the streaming setting and is designed to meet strict SLOs.

\paragraph{Parallel Streaming Inference Serving.}
Live streaming is critical not only for end users but, more importantly, for broadcasters and platforms operating cloud-scale services, where parallelism is essential to unlock capacity and quality. To accelerate diffusion models on multi-GPU systems~\citep{shih2023parallel, li2024distrifusion, fang2024xdit, fang2024pipefusion}, two complementary strategies—Sequence Parallelism (SP) and Pipeline Parallelism (PP)—are widely used. DeepSpeed-Ulysses~\citep{jacobs2023deepspeed}, Ring Attention~\citep{liu2023ring}, and their combination~\citep{fang2024xdit} realize sequence-parallel attention by sharding tokens across GPUs: Ulysses employs all-to-all communication to parallelize attention, whereas Ring Attention circulates key–value blocks in a ring topology to reduce communication. Distrifusion~\citep{li2024distrifusion} further introduces an asynchronous variant that overlaps the computation between spatiotemporal patches and timesteps. PipeFusion~\citep{fang2024pipefusion} applies pipeline parallelism with similar insights and achieves competitive efficiency. In our live-streaming framework, we find that barely utilizing either sequence-parallelism or pipeline parallelism leads to low FPS. We adopt a tailored parallelization scheme that explicitly balances compute and communication across heterogeneous hardware to meet strict latency and throughput targets.

\begin{figure}[t]
\centering
\begin{minipage}[t]{0.48\linewidth}
    \centering
    \includegraphics[width=\linewidth]{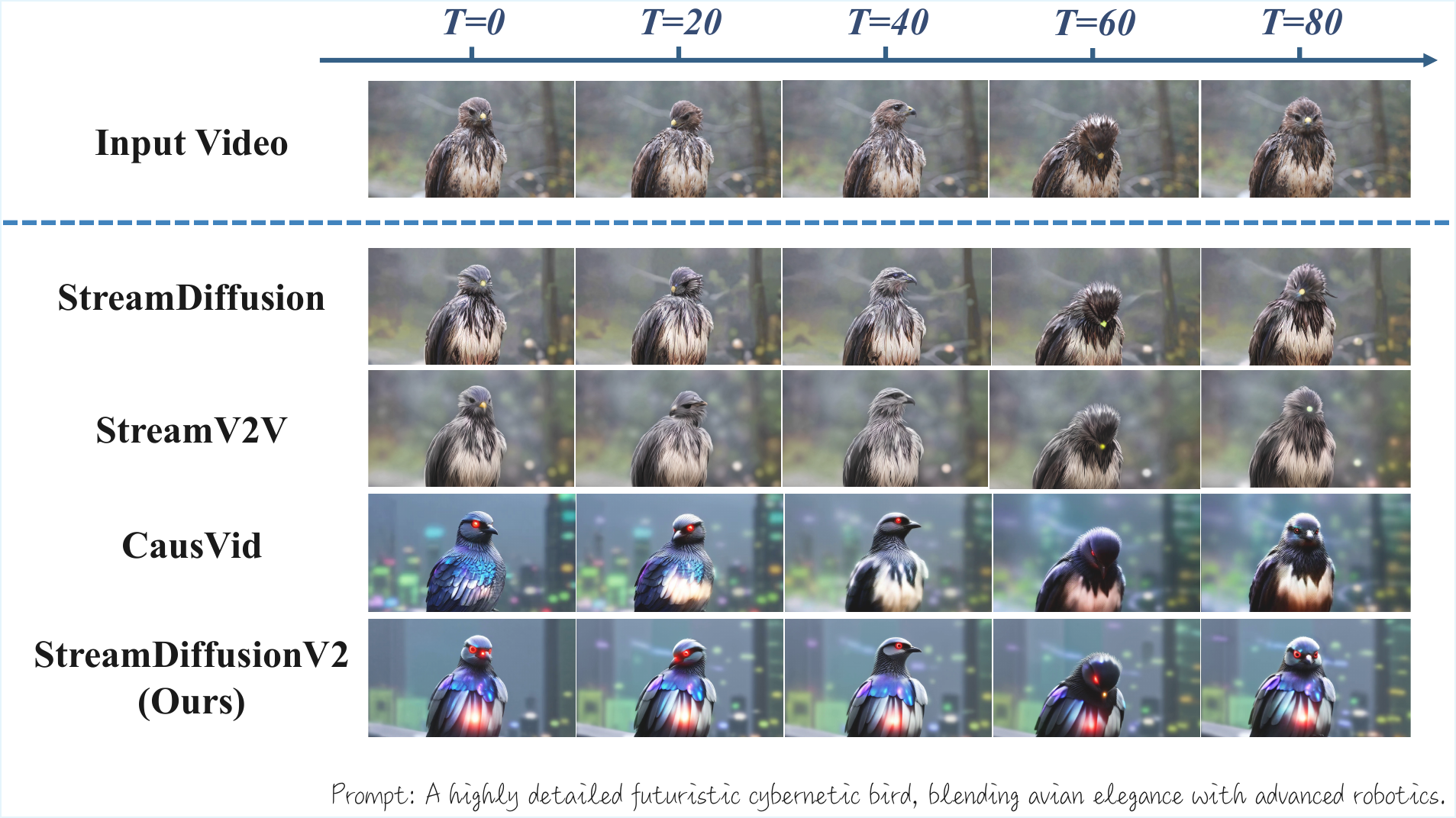}
    \vspace{-5pt}
    \caption{\textbf{Generation results among various approaches.} The examples above are frames picked from transferred videos among different methods, where the frame index is denoted as $T$.}
    \label{fig: prior_failure}
\end{minipage}
\hfill
\begin{minipage}[t]{0.48\linewidth}
    \centering
    \includegraphics[width=\linewidth]{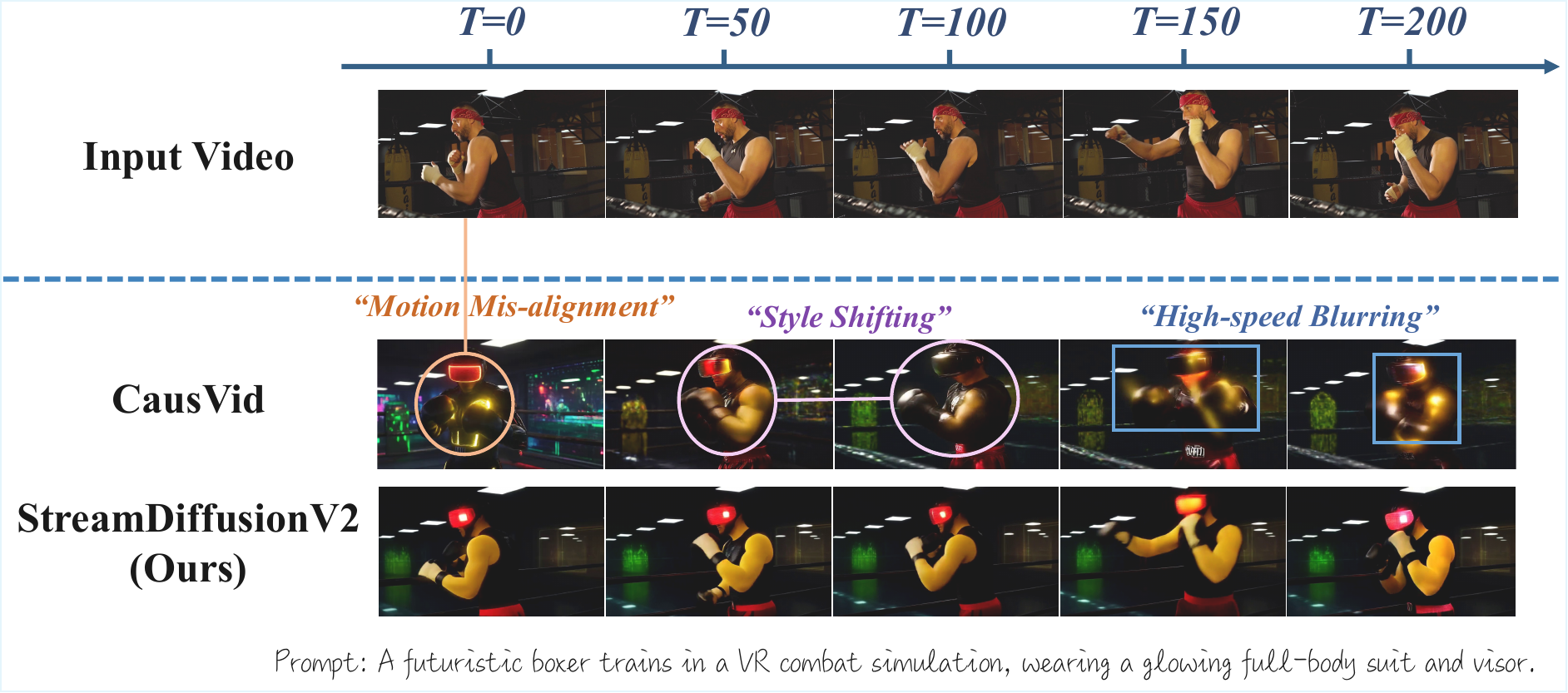}
    \vspace{-5pt}
    \caption{\textbf{Generation results of CausVid and ours.} The examples above are frames picked from transferred videos generated from CausVid and StreamDiffusionV2. We utilize the 1.3B model to produce the results.}
    \label{fig: causvid_failure}
\end{minipage}
\vspace{-8pt}
\end{figure}

\section{Motivation \& Bottleneck Analysis} \label{sec: motivation}

Real-time video applications span diverse use cases with widely varying budgets for frame rate, resolution, latency, and motion. This heterogeneity shifts performance bottlenecks across different stages of the pipeline. We highlight four key bottlenecks below.

\paragraph{Fixed-size input cannot satisfy real-time SLOs.}

Existing streaming systems adopt a fixed-input strategy, processing tens to hundreds of frames per forward pass to maximize throughput. For instance, CausVid and Self-Forcing process 81 frames per step.
While this large-chunk design improves average throughput in offline settings, it fundamentally conflicts with the requirements of real-time streaming.

Let $B$ denote batch size, $T$ the number of frames per forward pass, $(H, W)$ the video resolution, $P_{\text{model}}$ the model size (in parameters), $\rho_{\text{VAE}}$ the pixel-to-token ratio, and $C_{\text{device}}$ the average device throughput (FLOPs/s). Then, the time-to-first-frame (TTFF) can be approximated as:
\begin{equation}
    \text{TTFF} \approx \frac{2 \, B \, T \, H \, W \, P_{\text{model}}}{C_{\text{device}} \, \rho_{\text{VAE}}}.
\end{equation}
On a single H100 GPU, generating a 480p video with an 81-frame chunk using a 1.3B-parameter model yields a theoretical TTFF of 5.31s, which closely matches our measurements in Figure \ref{fig: ttff}. This delay far exceeds typical live-streaming targets ($\approx$1s\footnote{\url{https://www.mux.com/docs/guides/data-startup-time-metric}}).

The excessive latency is due to the fixed large input size, which does not adapt to real-time constraints. Without adaptive input scheduling, such pipelines cannot satisfy the latency requirements of live-streaming SLOs.

\paragraph{Drift accumulation in long-horizon generation.} 
\begin{wrapfigure}{r}{0.5\textwidth}
\vspace{-1cm}
\begin{center}
\includegraphics[width=0.48\textwidth]{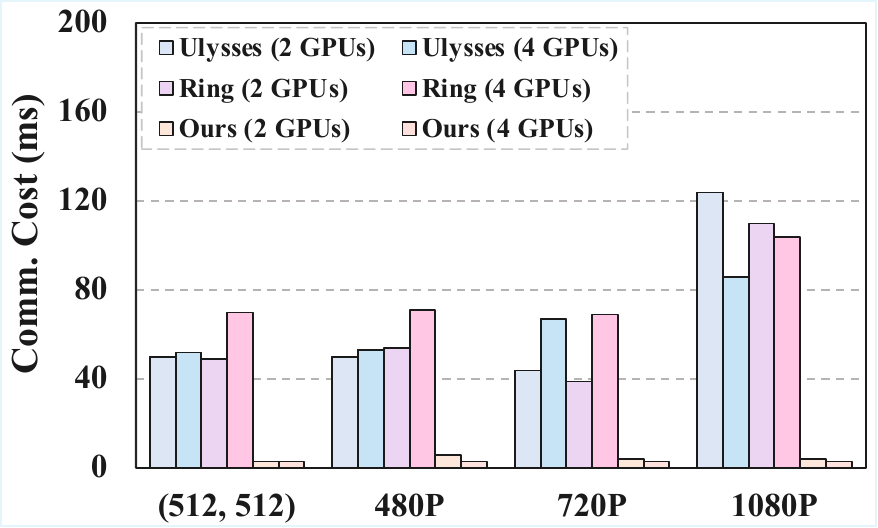}
\end{center}
\caption{\textbf{Communication consumption of various parallelism methods.} We measure the communication latency by testing the parallel inference latency and theoretical latency (sequence or block partitioning without communication) on NVlink-connected H100 GPUs.}
\vspace{-0.5cm}
\label{fig: com_cost}
\end{wrapfigure}

Current ``streaming'' video systems are primarily adapted from offline, bidirectional clip generators. For example,
CausVid \cite{yin2025slow} and Self-Forcinig \cite{huang2025self} are derived from Wan-2.1-T2V~\citep{wan2025wan}. These models are trained for short clips (5–10 seconds) and maintain coherence only within that range~\citep{huang2025self}.
Beyond this horizon, quality degrades rapidly (see Fig.~\ref{fig: prior_failure}), as the temporal context is treated \textit{statically}: the sink tokens become stale, RoPE accumulates positional drift, and the fixed context windows do not adapt to evolving content statistics. Over long durations, such compounding errors make these architectures inherently unsuitable for continuous live streaming.

\paragraph{Quality degradation due to motion unawareness.}

Different motion patterns in the input stream impose distinct tradeoffs between latency and visual quality. 
Fast motion requires conservative denoising to prevent tearing, ghosting, and blur, whereas slow or static scenes benefit from stronger refinement to recover details. 
Existing streaming pipelines rely on \textit{fixed} noise schedules that ignore this variability, leading to temporal artifacts in high-motion regions and reduced visual fidelity in low-motion segments (see Fig. \ref{fig: causvid_failure}).

\paragraph{Poor GPU scaling under per-frame latency constraints.}
In live-streaming scenarios, strict per-frame deadlines hinder the scalability of conventional parallelization strategies for two key reasons:
(i) communication latency in sequence parallelism significantly reduces potential speedup, and (ii) short-frame chunks drive the workload into a memory-bound regime. These effects are further amplified in real-time streaming, where efficient causal DiTs operate on short sequences (e.g., 4 frames per step), reducing per-frame computation and making communication overhead proportionally heavier (see Fig.~\ref{fig: com_cost}). We provide more information about the hardware intensity analysis and the configurations in Appendix Sec.~\ref{appendix_B}.

\section{Methodology}
In this section, we introduce StreamDiffusionV2, a training-free streaming system that achieves both real-time efficiency and long-horizon visual stability.
At a high level, our design is based on two key layers of optimization:
(1) \textbf{real-time scheduling and quality control}, which integrates SLO-aware batching, adaptive sink and RoPE refresh, and motion-aware noise scheduling to meet per-frame deadlines while maintaining long-horizon temporal coherence and visual fidelity; and
(2) \textbf{scalable pipeline orchestration}, which parallelizes the diffusion process across denoising steps and network stages to achieve near-linear FPS scaling without violating latency guarantees.
Additionally, we investigate several lightweight system-level optimizations, including DiT block scheduler, stream-VAE, and asynchronous communication overlap, which further enhance throughput and stability during long-running live streams.

\subsection{Real-time scheduling and quality control}

\begin{figure*}[t]
\centering
\includegraphics[width=\linewidth]{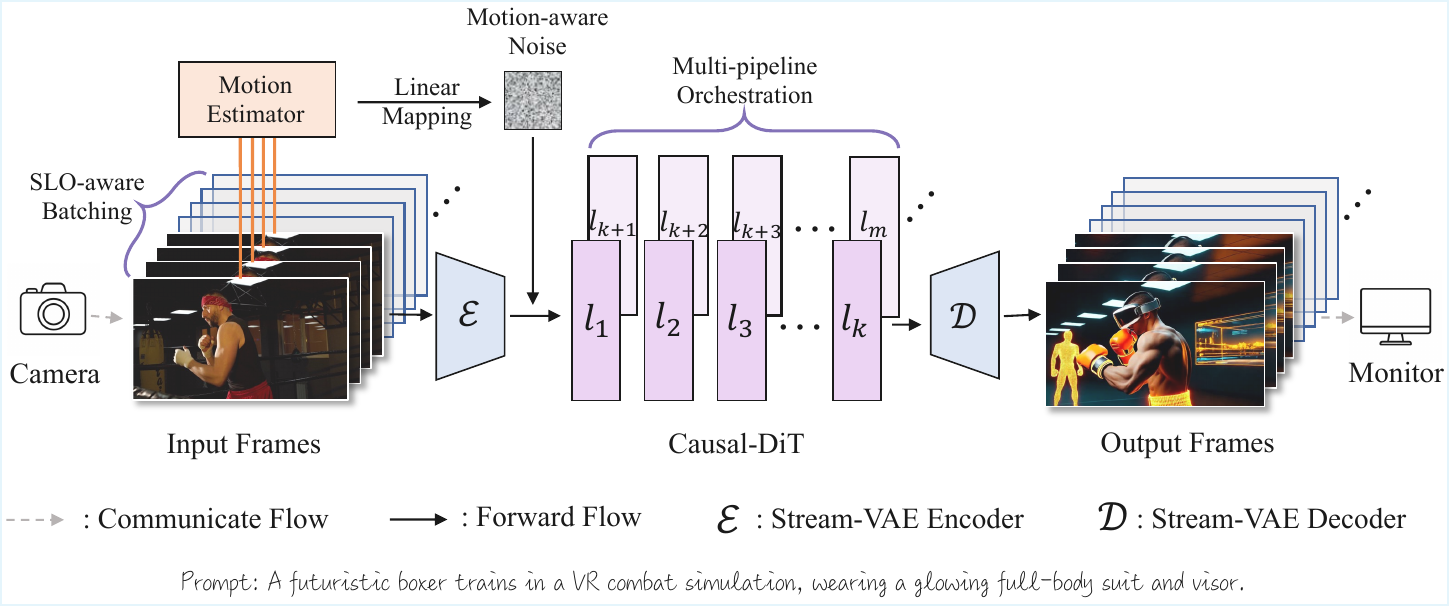}
\vspace{-15pt}
\caption{\textbf{The overview pipeline of our StreamDiffusionV2.} (1) \textbf{Efficiency.} We pair an SLO-aware batching scheduler (controlling input size) with a pipeline orchestration that balances latency and FPS, ensuring each frame meets its deadline and TTFF under strict service constraints.
(2) \textbf{Quality.} We deploy a motion-aware noise controller to mitigate high-speed tearing, and combine adaptive sink tokens with RoPE refreshing to deliver high-quality user interaction and hours-level streaming stability.}
\label{fig: pipeline}
\vspace{-10pt}
\end{figure*}

\begin{figure*}[t]
\centering
\includegraphics[width=\linewidth]{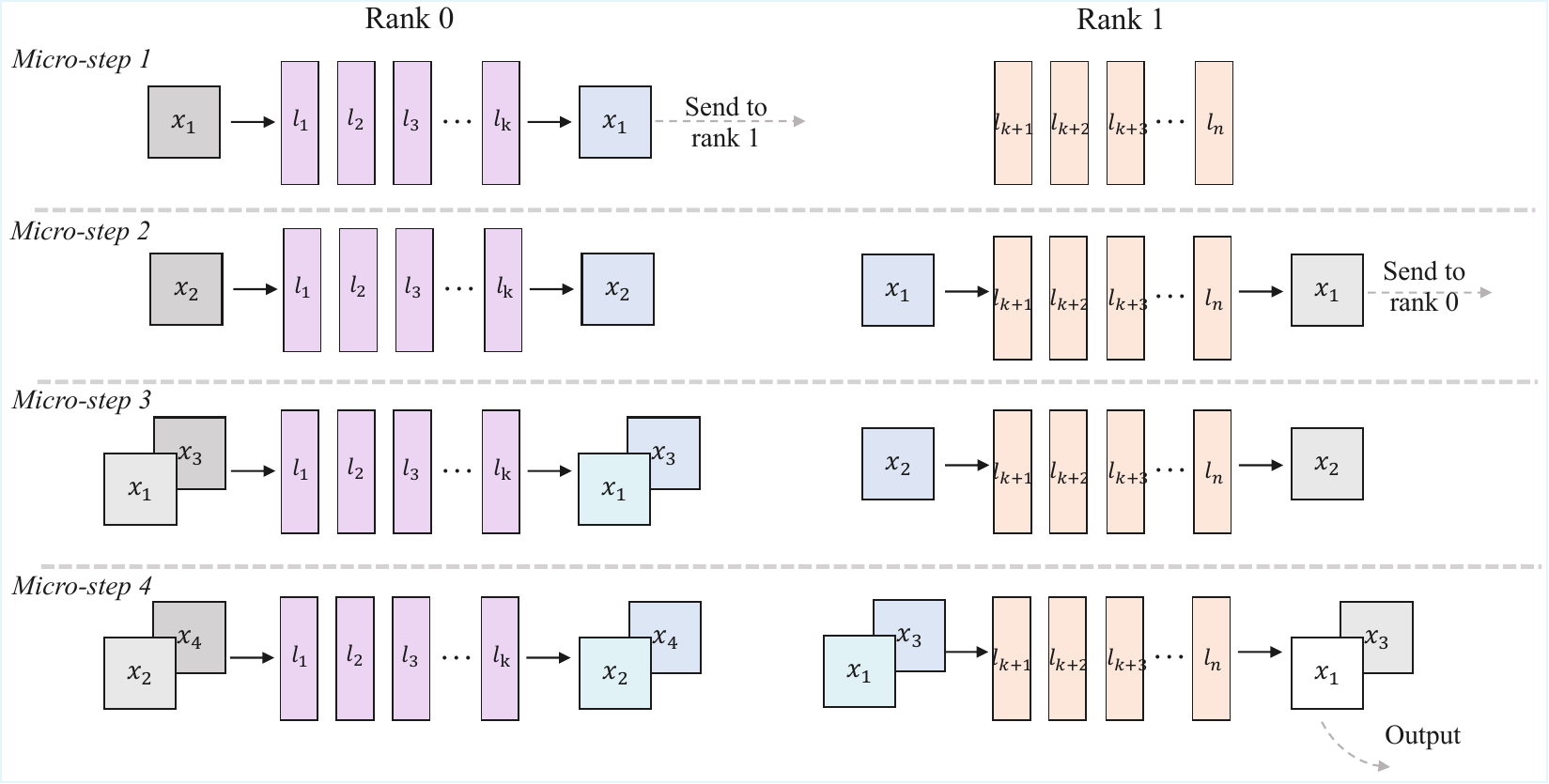}
\vspace{-10pt}
\caption{\textbf{The detailed design of our Pipeline-parallel Stream-Batch architecture.} The DiT blocks are distributed across multiple devices for pipeline parallelism, while the Stream-Batch strategy is applied within each stage. Different colors denote distinct latent streams, illustrating the communication structure, and the depth indicates the corresponding noise levels. Our implementation guarantees the generation of a clean latent at every micro-step during inference.}
\vspace{-10pt}
\label{fig: parallel}
\end{figure*}

As illustrated in Figure~\ref{fig: pipeline}, StreamDiffusionV2 achieves real-time video generation through three key components:
(1) an SLO-aware batching scheduler that dynamically adjusts the stream batch size to satisfy per-frame deadlines while maximizing GPU utilization;
(2) an adaptive sink and RoPE refresh mechanism that mitigates long-horizon drift by periodically resetting temporal anchors and positional offsets; and
(3) a motion-aware noise scheduler that adapts the denoising trajectory to motion magnitude, ensuring sharpness and temporal stability across diverse motion regimes.

\paragraph{SLO-aware batching scheduler.}
To satisfy the Service-Level Objective (SLO) while maximizing GPU utilization, we propose an SLO-aware batching scheduler that dynamically adjusts the batch size. 
Given a target frame rate $f_{\mathrm{SLO}}$, the system processes $T$ frames per iteration, with the overall inference latency depending on both the chunk size $T$ and batch size $B$, denoted as $L(T, B)$. 
To ensure real-time processing, the product $ B\cdot T$ must not exceed the number of frames already collected from the input stream. 
As analyzed in Section~\ref{sec: motivation}, the model operates in a memory-bound regime, and the inference latency can be approximated as:
\begin{equation}
L(T, B) \approx \frac{A(T, B) + P_{\mathrm{model}}}{\eta \, \mathrm{BW}_{\mathrm{HBM}}},
\end{equation}
where $A(T, B)$ denotes the activation memory footprint, $P_{\mathrm{model}}$ represents the memory volume of the model parameters, and $\eta \, \mathrm{BW}_{\mathrm{HBM}}$ is the effective memory bandwidth with the utilization factor $\eta$ $(0 < \eta \le 1)$. 
With FlashAttention~\citep{dao2022flashattention}, the activation term $A(T, B)$ scales linearly as $\mathcal{O}(BT)$, resulting in a proportional latency growth $L(T, B)$.
The achieved processing frequency can therefore be expressed as $f = B T / L(T, B) \propto \frac{B}{1 + B}$, which increases with larger batch size $B$ as GPU utilization improves.
As the system approaches the knee point of the roofline model (Fig.~\ref{fig: roofline})—marking the transition from the memory-bound to the compute-bound regime—the scheduler adaptively converges to an optimal batch size $B^\ast$ that maximizes throughput efficiency.

\paragraph{Adaptive sink and RoPE refresh.}
To address the \textit{drifting issue} discussed in Section~\ref{sec: motivation}, we introduce an adaptive sink token update and a RoPE refresh policy that jointly maintain long-horizon stability during continuous video generation.
Unlike prior methods such as Self-Forcing~\citep{huang2025self}, which fix the sink set throughout generation, StreamDiffusionV2 dynamically updates the sink tokens based on the evolving prompt semantics.
Let $\mathcal{S}_t = \{s_1^t, \ldots, s_m^t\}$ denote the sink set at chunk $t$. Given a new chunk embedding $\mathbf{h}_t$, the system computes similarity scores $\alpha_i = \cos(\mathbf{h}_t, s_i^{t-1})$ and refreshes the least similar sinks: $s_i^{t} = s_i^{t-1}$ if $\alpha_i \ge \tau$, and $s_i^{t} = \mathbf{h}_t$ otherwise, where $\tau$ is a similarity threshold. In practice, we find that $\tau$ should be set large to ensure alignment with the evolved text.
To prevent positional drift caused by accumulated RoPE offsets over long sequences, we periodically reset the RoPE phase once the current frame index $t$ exceeds a threshold $T_{\text{reset}}$, i.e., $\theta_t = \theta_t$ if $t \le T_{\text{reset}}$ and $\theta_t = \theta_{t - T_{\text{reset}}}$ otherwise.

\paragraph{Motion-aware noise scheduler.}

\begin{wrapfigure}{r}{0.5\textwidth}
\vspace{-0.4cm}
\begin{center}
\includegraphics[width=0.5\textwidth]{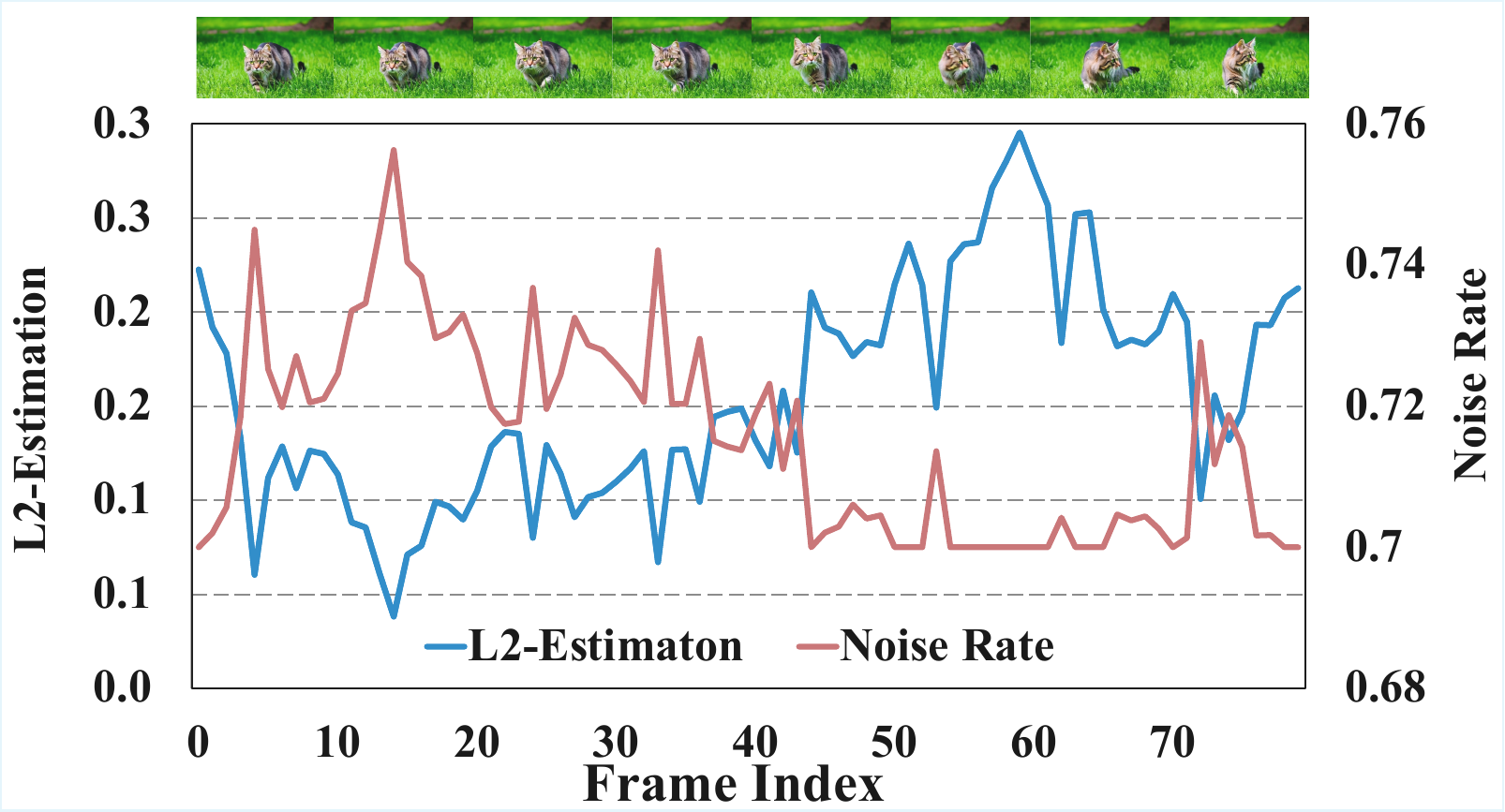}
\end{center}
\caption{\textbf{Example of motion estimation and dynamic noise rate.} The curves indicate the L2-estimation and its corresponding noise rate of the video.}
\vspace{-0.5cm}
\label{fig: motion}
\end{wrapfigure}

To handle diverse motion dynamics in live-streaming videos, we propose a motion-aware noise scheduler that adaptively regulates the denoising noise rate according to the estimated motion magnitude of recent frames.
As illustrated in Fig.~\ref{fig: motion}, we estimate the motion magnitude between consecutive frames using a frame-difference metric. Given consecutive latent frames $\mathbf{v}_t, \mathbf{v}_{t-1} \in \mathbb{R}^{C \times H \times W}$, the motion intensity $d_t$ is 
\[
d_t = \sqrt{\frac{1}{CHW}\|\mathbf{v}_t - \mathbf{v}_{t-1}\|_2^2}.
\]
To stabilize this measurement over a short temporal window of $k$ frames, we normalize it by a statistical scale factor $\sigma$ and clip it into $[0, 1]$:
\[
\hat{d}_t = \mathrm{clip}\!\left(\frac{1}{\sigma}\max_{i \in \{t-k,\ldots,t\}} d_i,\, 0,\, 1\right).
\]
The normalized $\hat{d}_t$ determines how aggressively the system should denoise the current chunk. A higher $\hat{d}_t$ (fast motion) corresponds to a more conservative denoising schedule, while a lower $\hat{d}_t$ (slow or static motion) allows stronger refinement for sharper details. 
Finally, we smooth the noise rate $s_t$ using an exponential moving average (EMA) to ensure gradual temporal transitions:  
\[
s_t = \lambda \left[s_{\max} - (s_{\max} - s_{\min})\hat{d}_t\right] + (1-\lambda)s_{t-1},
\]
where $0 < \lambda < 1$ controls the update rate, and $s_{\max}$ and $s_{\min}$ denote the upper and lower bounds of the noise rate.

\subsection{Scalable pipeline orchestration}
\paragraph{Multi-Pipeline orchestration scaling.}
To improve system throughput on multi-GPU platforms, we propose a scalable pipeline orchestration for parallel inference. Specifically, the DiT blocks are partitioned across devices. As illustrated in Fig.~\ref{fig: parallel}, each device processes its input sequence as a micro-step and transmits the results to the next stage within a ring structure. These enable consecutive stages of the model to operate concurrently in a pipeline-parallel manner, achieving near-linear acceleration for DiT throughput. 

Notably, the pipeline-parallel inference adds inter-stage communication, which, together with activation traffic, keeps the workload memory-bound. 
To cope with this and still meet real-time constraints, we extend the SLO-aware batching mechanism to the multi-pipeline setting and combine it with the batch-denoising strategy. Concretely, we produce a fine-denoised output at every micro-step (Fig.~\ref{fig: parallel}), while treating the $n$ denoising steps as an effective batch multiplier, yielding a refined latency model $L(T, nB)$. The scheduler continuously adapts $B$ to the observed end-to-end latency so that the per-stream rate satisfies $f_{\mathrm{SLO}}$, while the aggregate throughput approaches the bandwidth roofline.

\subsection{Efficient system-algorithm co-design}

\paragraph{DiT Block Scheduler.}
Static partitioning often produces unbalanced workloads because the first and last ranks handle VAE encoding and decoding in addition to DiT blocks, as shown in Fig.\ref{fig: balancing}(a). This imbalance leads to pipeline stalls and reduced utilization~\citep{yeung2024balancing}.
We introduce a lightweight inference-time \textit{DiT block scheduler} that dynamically reallocates blocks between devices based on measured execution time. The scheduler searches for an optimal partition that minimizes per-stage latency, as illustrated in Fig.~\ref{fig: balancing}(b), significantly reducing overall pipeline bubbles.

\paragraph{Stream-VAE.}
StreamDiffusionV2 integrates a low-latency Video-VAE variant designed for streaming inference. Instead of encoding long sequences, Stream-VAE processes short video chunks (e.g., 4 frames) and caches intermediate features within each 3D convolution to maintain temporal coherence.

\paragraph{Asynchronous communication overlap.}
To further reduce synchronization stalls, each GPU maintains two CUDA streams: a computation stream and a communication stream. Inter-GPU transfers are executed asynchronously, overlapping with local computation to hide communication latency.
This double-stream design aligns each device’s compute pace with its communication bandwidth, effectively mitigating residual bubbles and sustaining high utilization across the multi-GPU pipeline.

\section{Experiments}

\subsection{Setup}
\paragraph{Models.} 
The StreamDiffusionV2 model is built on Wan 2.1~\citep{wan2025wan} and CausVid~\citep{yin2025slow}. The proposed method is training-free. In the Appendix, we describe how we can improve the visual quality by efficient finetuning with REPA~\citep{yu2024repa} to cater to the user applications. 

\paragraph{Efficiency metrics.}
We benchmark throughput under varying methods and configurations. For delay analysis, we report FPS (frames per second) and time-to-First-Frame (TTFF) for end-to-end latency. We also report the acceleration rate = (baseline inference time) / (optimized inference time), which directly quantifies overall speedup.

\paragraph{Quality metrics.}
Following prior work~\citep{liang2024looking}, we compute the \textbf{CLIP Score} as the cosine similarity between the CLIP~\citep{radford2021learning} features of the generated and reference frames. We further measure \textbf{Warp Error} by estimating optical flow between consecutive inputs with RAFT~\citep{teed2020raft} and warping the corresponding generated frames; lower is better. These metrics capture semantic- and pixel-level consistency, respectively.
\paragraph{Baselines.}
For efficiency, we compare with sequence-parallel approaches: Ring-Attention~\citep{liu2023ring} and DeepSpeed-Ulysses~\citep{jacobs2023deepspeed}. For generation quality, we include StreamDiffusion~\citep{kodaira2023streamdiffusion}, StreamV2V~\citep{liang2024looking}, and a video-to-video variant of CausVid (implemented via a naïve noising–denoising scheme).

\paragraph{Implementation details.}
We evaluate StreamDiffusionV2 on enterprise- and consumer-grade GPUs—$4\times$ H100 (80\,GB, NVLink) and $4\times$ RTX~4090 (24\,GB, PCIe).
All runs use \texttt{bf16} with no TensorRT or quantization.
Results are reported at $512\times512$ and 480p ($832\times480$) with $1$–$4$ denoising steps.

\subsection{Efficiency Evaluation}

\begin{figure*}[t]
\centering
\includegraphics[width=\linewidth]{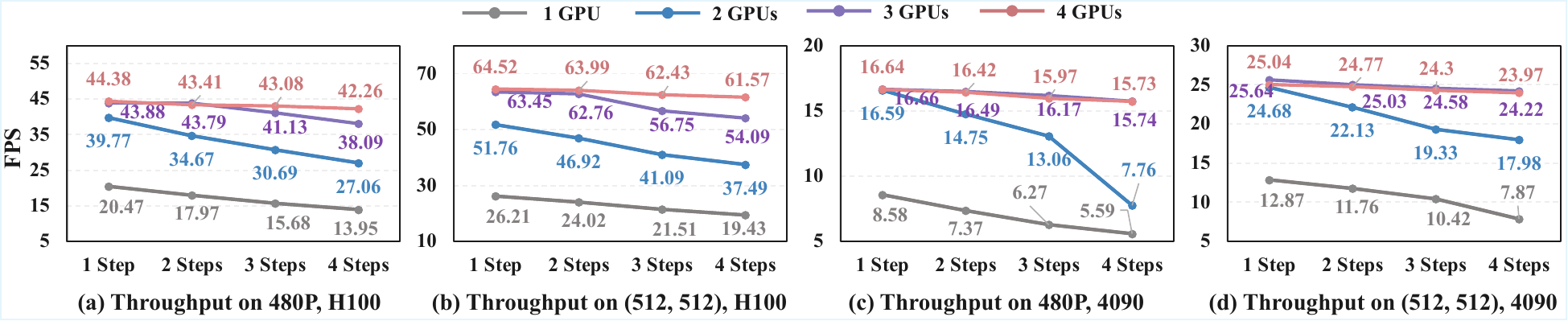}
\vspace{-15pt}
\caption{\textbf{The throughput results of the 1.3B model on H100 GPUs (with NVLink) and 4090 GPUs (with PCIe)} among different denoising steps and various resolutions. We report the result without batching on 4090 because of the memory limitation.}
\vspace{-10pt}
\label{fig: fps}
\end{figure*}

\begin{figure}[t]
\centering
\begin{minipage}[t]{0.48\linewidth}
    \centering

    \includegraphics[width=\linewidth]{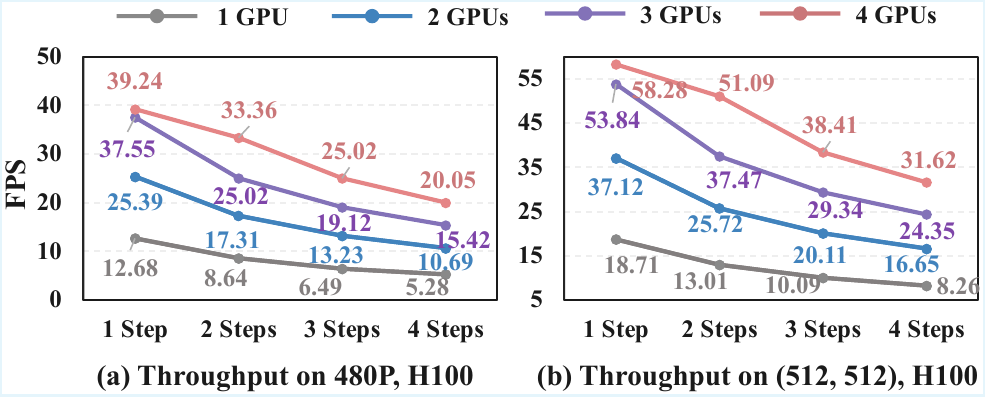}
    \vspace{-5pt}
    \caption{\textbf{The throughput results of the 14B model on H100 GPUs (communicate through NVLink)} among different denoising steps and various resolutions. }
    \label{fig: fps_14b}
\end{minipage}
\hfill
\begin{minipage}[t]{0.48\linewidth}
    \centering
    \includegraphics[width=\linewidth]{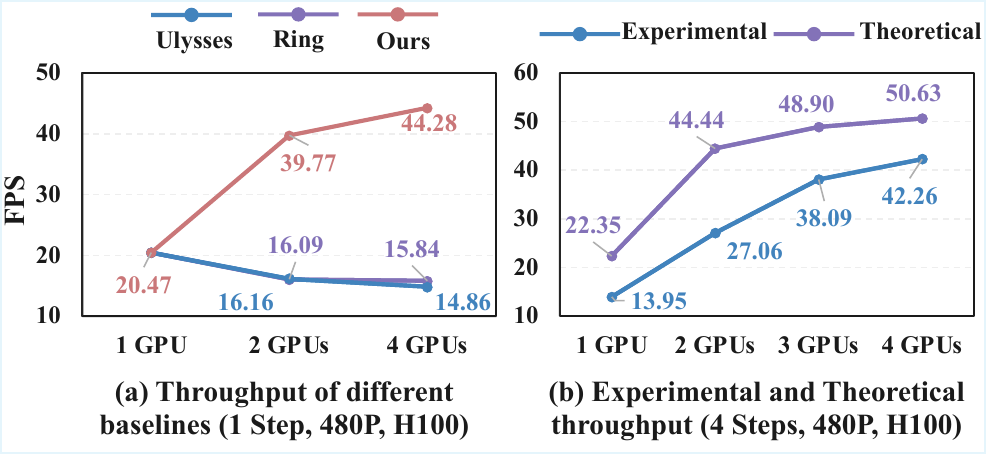}
    \vspace{-5pt}
    \caption{(a) The throughput results of different parallel inference systems on H100 GPUs. (b) The theoretical throughput of the proposed system and experimental results.}
    \label{fig: fps_baselines_theor}
\end{minipage}
\vspace{-8pt}
\end{figure}

\begin{figure}[t]
\centering
\begin{minipage}[t]{0.48\linewidth}
    \centering
    \includegraphics[width=\linewidth]{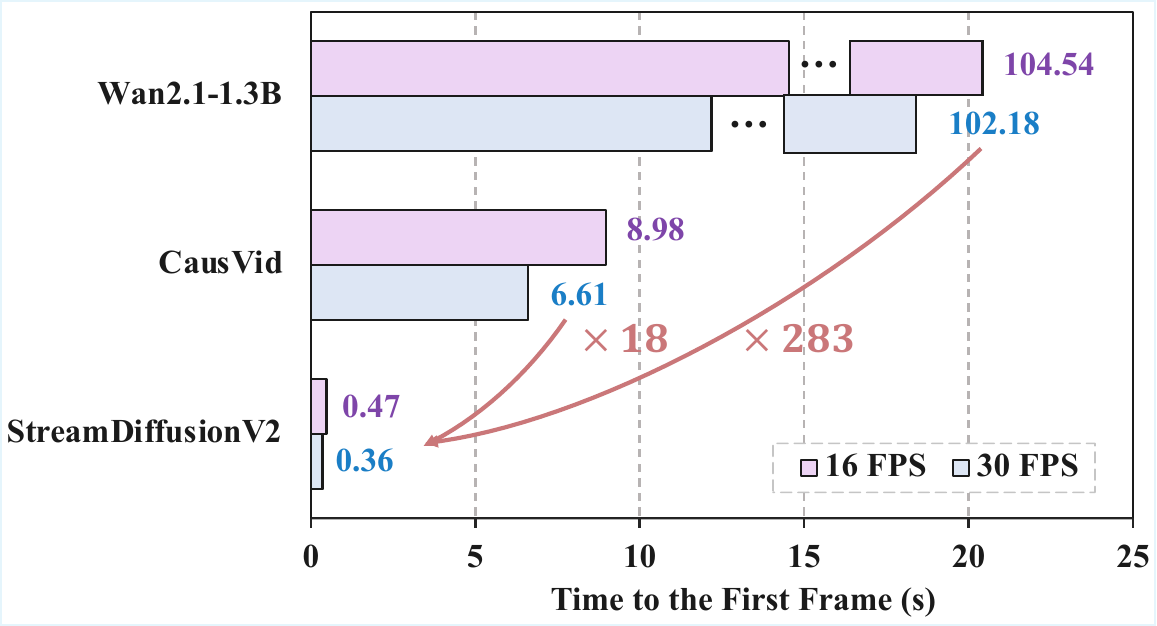}
    \vspace{-5pt}
    \caption{\textbf{Time to the first frame on H100 GPU.} We present the 2-step denoising results of CausVid and StreamDiffusionV2, 50 steps denoising results on Wan2.1-T2V-1.3B.}
    \label{fig: ttff}
\end{minipage}
\hfill
\begin{minipage}[t]{0.48\linewidth}
    \centering
    \includegraphics[width=\linewidth]{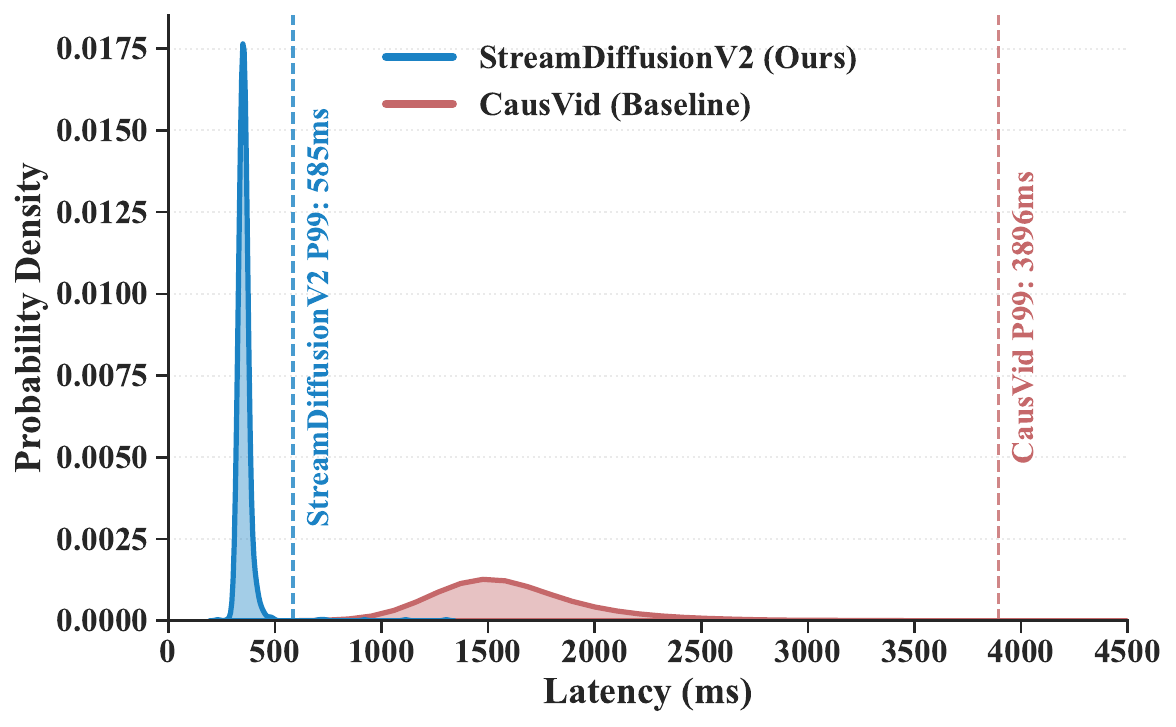}
    \vspace{-5pt}
    \caption{\textbf{End-to-end latency distributions of StreamDiffusionV2 and the baseline.}}
    \label{fig: latency}
\end{minipage}
\vspace{-8pt}
\end{figure}

\subsubsection{TTFF Results}

We compare the Time-to-First-Frame (TTFF) results of Wan-T2V-1.3B, CausVid, and our proposed StreamDiffusionV2 during video-to-video generation. The time-to-first-frame (TTFF) depends on the configured input/output FPS and frame chunk size, and is calculated through the sum of the frame buffering delay and the processing latency. Taking advantage of the SLO-aware input and streaming VAE design, StreamDiffusionV2 achieves a substantial reduction in TTFF, reaching 0.47s and 0.37s at 16 FPS and 30 FPS video throughput, respectively. Specifically, at 30 FPS, CausVid and Wan2.1-1.3B exhibit 18× and 280× higher TTFF than our pipeline, respectively. The TTFF metric indicates the actual latency in a live-streaming application, which has demonstrated the capacity of the proposed pipeline in interactive real-time generation.

\subsubsection{FPS Results}

Fig.~\ref{fig: fps} presents the speed under different resolutions and GPU configurations, with the Wan-T2V-1.3B models. On the H100 platform, which benefits from high-bandwidth NVLink interconnects, StreamDiffusionV2 achieves 42.26 FPS at 480P and 61.57 FPS at 512$\times$512 with a 1-step model, as shown in Fig.~\ref{fig: fps} (a) and (b), respectively. Even when the denoising steps increase to four, the system still produces more than 40 FPS at 480P and 60 FPS at 512$\times$512, showing stable performance under heavier diffusion workloads. Moving toward custom devices such as 4090 GPUs with PCIe connections, we still achieve nearly 16 FPS in 480P and 24 FPS in 512$\times$512, respectively.

Moreover, we evaluate our method on a 14B parameter configuration to assess scalability for large diffusion backbones, as shown in Fig.~\ref{fig: fps_14b}. Despite the substantial model size, the proposed Pipeline-parallel Stream-batch design achieves 39.24 FPS at 480P and 58.28 FPS at 512$\times$512 across 4 GPUs, showing that the system remains compute-efficient and communication-balanced under heavy workloads. In particular, our pipeline achieves comparable throughput on the 14B-parameter model. This is mainly because both the 1.3B and 14B models share the same VAE weights, while the VAE accounts for approximately 30\% of the total inference time. As a result, the VAE’s processing time remains constant and the increased computational cost only impacts the DiT component. The time consumption caused by VAE also leads to a deviation between the ideal and actual throughput gains for the whole pipeline.

\subsubsection{Throughput Analysis}
\paragraph{Baseline throughput.}
We report throughput results of Deepspeed-Ulysses~\cite{jacobs2023deepspeed} and Ring-Attention~\cite{liu2023ring} for comparison. Due to the communication overhead introduced by sequence parallelism, neither of these methods has achieved the throughput gain. A detailed result of the throughput gain is provided in Fig.~\ref{fig: sp_inference}, Appendix~\ref{appendix_D}.

\paragraph{Theoretical throughput.}
Following our performance-regime analysis and based on profiling, we conservatively assume a 50\% rate bubble, and use the isolated VAE runtime $t_{\text{VAE}}$ as the ideal VAE inference time. The theoretical latency $t_{\text{theory}}$ is calculated as:
\begin{equation}
    t_{\text{theory}} = \frac{B / \text{bandwidth}}{\text{rate\_bubble}} + t_{\text{VAE}}.
\end{equation}

The theoretical throughput is given by
\begin{equation}
    \text{FPS}_{\text{theory}} = \frac{\text{chunk\_size}}{t_{\text{theory}}},
\end{equation}
and the results are shown in Fig.~\ref{fig: fps_baselines_theor}(b). The experimentally measured throughput closely follows the same scaling trend as the theoretical throughput, and approaches the theoretical values under certain configurations.

\subsubsection{SLO-oriented Metrics}
\begin{table}[t]
\centering
\caption{Comparison of Latency, Jitter, and SLO Miss Rate between StreamDiffusionV2 and Baseline (CausVid with StreamVAE) on an H100 Workstation.}
\vspace{4pt}
\label{tab: latency_comparison}
\small
\setlength{\tabcolsep}{6pt}
\renewcommand{\arraystretch}{1.15}

\begin{tabular}{llccccccc}
\toprule
\textbf{Method} & \textbf{Metric (ms)} 
& \textbf{Mean} & \textbf{Std.} 
& \textbf{P50} & \textbf{P90} & \textbf{P95} & \textbf{P99} 
& \textbf{Miss Rate (1s)} \\ 
\midrule

\multirow{2}{*}{StreamDiffusionV2 \textbf{(Ours)}} & Tail Latency & 357 & \textemdash & 361 & 484 & 497 & 585 & 0.2\% \\ 
& Jitter  & 21  & 30 & 13 & 45 & 61 & 132 & \textemdash \\
\midrule

\multirow{2}{*}{CausVid}
& Tail Latency 
& 1760 & \textemdash & 1672 & 2958 & 3304 & 3896 & 99.9\% \\
& Jitter        
& 235  & 255 & 164 & 498 & 599 & 1310 & \textemdash \\

\bottomrule
\end{tabular}
\end{table}

We report SLO-style tail latency together with jitter and miss rate metrics. Following common industry practice for interactive/live streaming\footnote{\url{https://www.mux.com/blog/the-low-latency-live-streaming-landscape-in-2019?}}, we target an end-to-end latency budget of 1.0s.

We provide the end-to-end latency distribution charts for our method and the baseline. As shown in Fig.~\ref{fig: latency}, the baseline model, CausVid, has a dispersed distribution, and most samples have missed the 1s SLO deadline. These will also lead to severe jitter and lag during real-time applications. In contrast, our StreamDiffusionV2 achieves a concentrated distribution to meet the SLO.

For quantitative results, as shown in Tab.~\ref{tab: latency_comparison}, on an H100 workstation, our method provides clear advantages in latency, stability, and SLO miss rate compared to the baseline. In particular, under a 1s SLO, the miss rate is only 0.2\%, indicating that StreamDiffusionV2 reliably meets the latency budget. For stability, the jitter is 21 ms (mean) with 30 ms standard deviation, suggesting stable frame delivery under steady-state serving. All results are obtained from the online video-to-video experiments, conducted with a single-step configuration at a resolution of 512×512 on an H100 GPU.

\subsection{Generation Quality Evaluation}
\label{sec: quality_comparison}

\begin{table}[t]
\centering
\small
\renewcommand{\arraystretch}{1.0}
\setlength{\tabcolsep}{3pt}  
\caption{\textbf{Quantitative metrics comparison.} We report the CLIP consistency and prompt score, and warp error to indicate the consistency of generated videos.}
\label{tab: quality_metrics}
\begin{tabular}{lcccc}
\toprule
 & StreamDiffusion & StreamV2V & CausVid & StreamDiffusionV2 \\
\midrule
 Text-Image CLIP~$\uparrow$ & 26.48 & - & 27.69 & \textbf{29.29} \\
 Temporal CLIP~$\uparrow$ & 95.24 & 96.58 & 98.48 & \textbf{98.51} \\
 Warp Error~$\downarrow$ & 117.01 & 102.99 & 78.71 & \textbf{73.31} \\
\bottomrule
\end{tabular}
\vspace{-5pt}
\end{table}

\begin{table}[t]
\centering
\small
\renewcommand{\arraystretch}{1.0}
\setlength{\tabcolsep}{3pt}       
\caption{\textbf{Quantitative metrics for ablation studies.} We report the CLIP score and warp error for ablation studies. The $\checkmark$ indicates adding the corresponding module to the baseline models.}
\vspace{5pt}
\label{tab: ablation_metrics}
\begin{tabular}{ccccc}
\toprule
 Sink Token & Dynamic Noising &  CLIP Score~$\uparrow$ &  Warp Error~$\downarrow$ \\
\midrule
 - & - & 98.38 & 79.51 \\
 - & $\checkmark$ & 98.36 & 75.71 \\
 $\checkmark$ & - & 98.47 & 73.64 \\
  $\checkmark$ & $\checkmark$ & \textbf{98.51} & \textbf{73.13} \\
\bottomrule
\vspace{-10pt}
\end{tabular}
\end{table}

\subsubsection{Comparison of Video Quality Metrics}
As shown in Tab.~\ref{tab: quality_metrics}, approaches based on image diffusion models, such as StreamDiffusion~\citep{kodaira2023streamdiffusion} and StreamV2V~\citep{liang2024looking}, exhibit noticeable temporal inconsistency, resulting in lower CLIP scores and higher Warp Errors. For CausVid~\citep{yin2025slow}, we implement a naive video-to-video generation baseline for fair quality evaluation. It achieves a comparable CLIP score but exhibits a higher Warp Error than our proposed method. These results indicate that our style-preserving and motion-aware strategies effectively enhance pixel-level temporal consistency, while maintaining comparable semantic similarity, as both methods share the same model weights.

\subsubsection{Comparison of Generation Results}
\begin{wrapfigure}{r}{0.5\textwidth}
\vspace{-0.5cm}
\begin{center}
\includegraphics[width=0.48\textwidth]{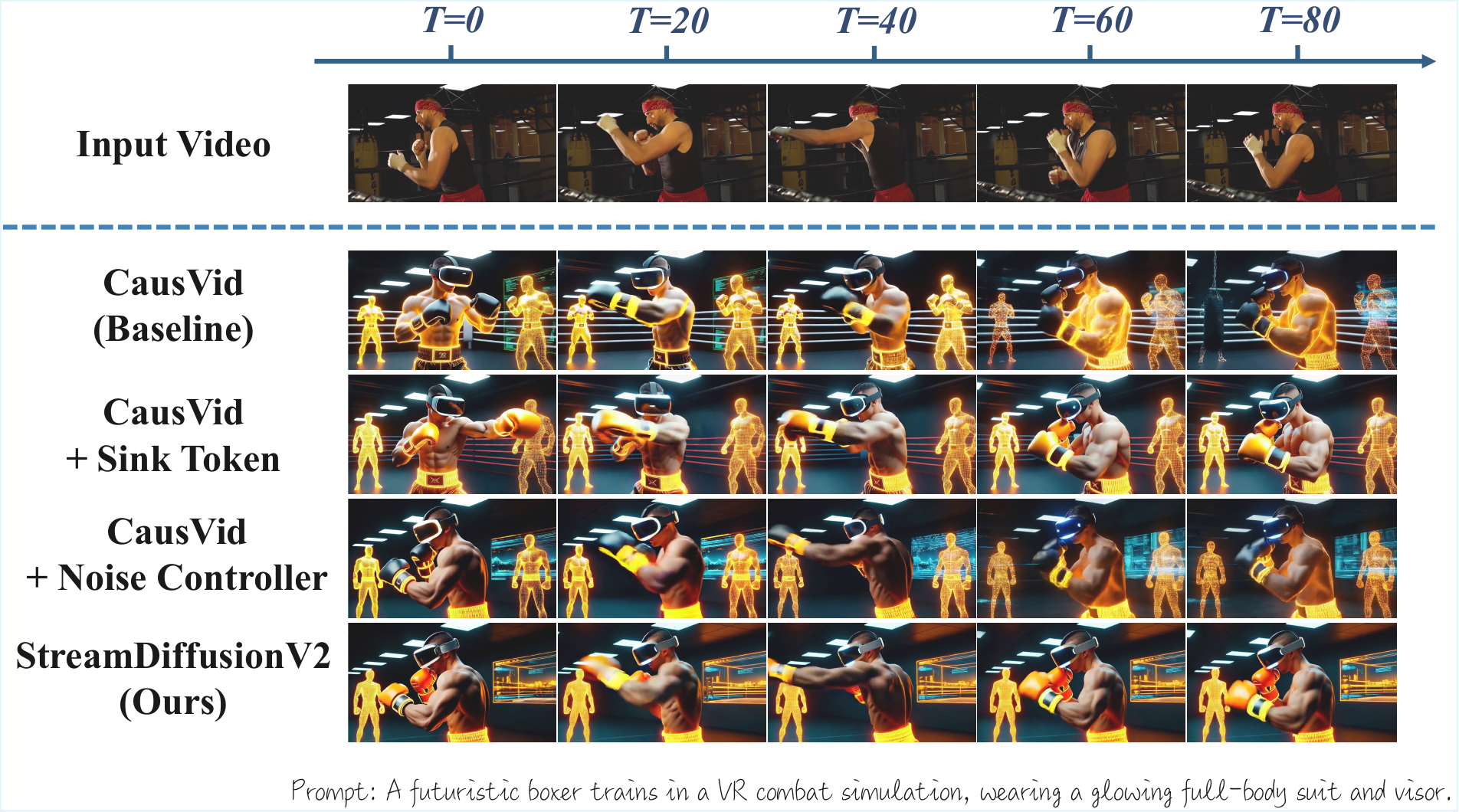}
\end{center}
\caption{\textbf{High-speed input video generation results across various configurations from 14B models.}}
\vspace{-0.4cm}
\label{fig: ablation_vis}
\end{wrapfigure}
We also present the generation results in Fig.~\ref{fig: prior_failure} for visualization comparison. Compared to previous approaches, which are based on image diffusion models, the proposed method achieves superior video transfer capabilities, consistent motion, and a broader range of results. Moreover, without sink tokens for style preservation, CausVid exhibits significant style fading as generation progresses. When input videos contain high-speed motion, as shown in Fig.~\ref{fig: ablation_vis}, CausVid produces motion-misaligned transferred frames, because of the over-smooth training data. In contrast, our proposed method leverages the sink token strategy to maintain visual style and employs a motion-aware noise controller for dynamic noising, resulting in a temporally consistent appearance and accurate motion structure.

\subsection{Analysis of Efficiency and Generation Quality}

\subsubsection{Effectiveness of Sink Token and Motion-Aware Noise Controller}

Following Sec.~\ref{sec: quality_comparison}, we evaluate the proposed modules using \textbf{CLIP Score} and \textbf{Warp Error}.
As reported in Table~\ref{tab: quality_metrics}, augmenting the baseline (live-stream–customized CausVid) with the \emph{Motion-Aware Noise Controller} slightly reduces CLIP Score but noticeably improves Warp Error.
This matches intuition: the controller adaptively lowers noise intensity in proportion to motion frequency, trading a small amount of semantic consistency for better pixel-level alignment.
When combined with the Sink Token, the pipeline achieves state-of-the-art performance on both metrics.

To disentangle the contributions, Fig.~\ref{fig: ablation_vis} shows a high-speed motion example where vanilla CausVid exhibits severe degradation.
The \emph{Sink Token} stabilizes global style across frames (see the target character and background), while the \emph{Motion-Aware Noise Controller} preserves motion structure between generated and reference frames, mitigating temporal misalignment.

\subsubsection{Effectiveness of the Dynamic DiT-Block Scheduler}
\begin{figure}[t]
\centering
\begin{minipage}[t]{0.48\linewidth}
    \centering
    \includegraphics[width=\linewidth]{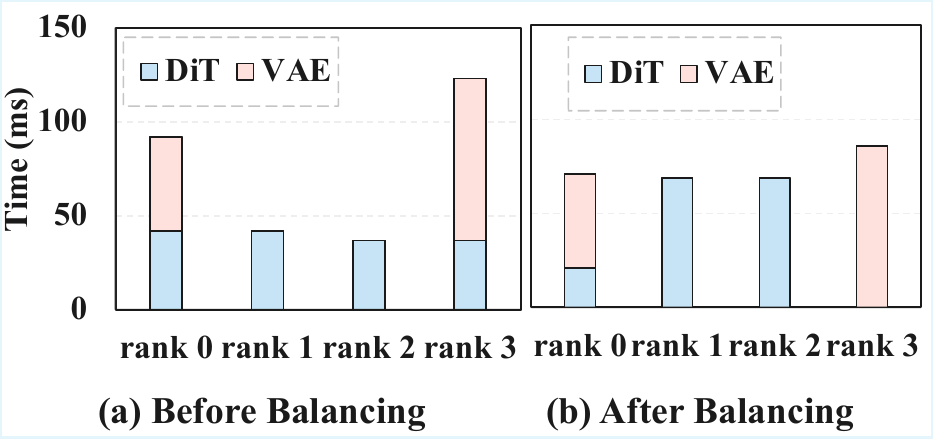}
    \vspace{-5pt}
    \caption{\textbf{Time consumption before and after the balancing schedule.} (a) Time consumption among various devices before balancing. (b) Time consumption after balancing. We present the 4-step denoising results on NVLink-connected H100 GPUs.}
    \label{fig: balancing}
\end{minipage}
\hfill
\begin{minipage}[t]{0.48\linewidth}
    \centering
    \includegraphics[width=\linewidth]{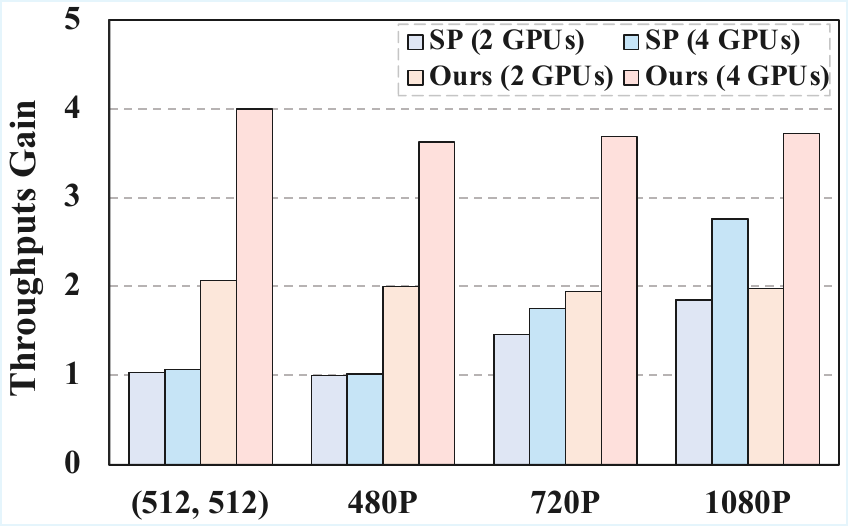}
    \vspace{-5pt}
    \caption{\textbf{Theoretical computation consumption of various parallelism acceleration methods on H100.} We test the time consumption of sequence partition and block partition to simulate the corresponding parallel inference approach.}
    \label{fig: theor_cost}
\end{minipage}
\vspace{-8pt}
\end{figure}

Balanced workload is critical for efficient pipeline-parallel inference.
We profile DiT vs.\ VAE cost and measure latency at 480p resolution with 4 denoising steps on a 1.3B model.
Figure~\ref{fig: balancing}(a) shows that Video VAE (our streaming implementation following the baseline designs~\citep{wan2025wan}) contributes substantially to the runtime and can induce stage imbalance.
Our \emph{dynamic scheduler} partitions DiT blocks to equalize per-device time, markedly reducing imbalance and improving throughput; see Fig.~\ref{fig: balancing}(b).

\subsubsection{Sequence Parallelism vs. Pipeline Orchestration}

Sequence parallelism is a widely used efficiency technique. Here we compare it with our pipeline orchestration along two axes: (i) \emph{communication cost} , and (ii) the \emph{performance-bound regime}.

\paragraph{Communication cost.}
\begin{wrapfigure}{r}{0.5\textwidth}
\vspace{-0.5cm}
\begin{center}
\includegraphics[width=0.5\textwidth]{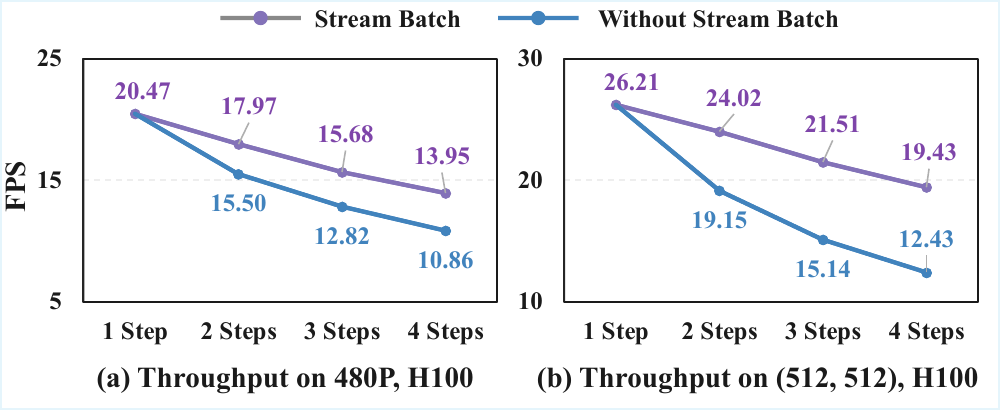}
\end{center}
\caption{\textbf{The throughput comparison between with and without Stream Batch.} }
\vspace{-0.5cm}
\label{fig: streambatch}
\end{wrapfigure}

As shown in Fig.~\ref{fig: com_cost}, we measure communication overhead by subtracting the ideal single-device latency from the observed distributed latency.
Across resolutions, both DeepSpeed-Ulysses~\citep{jacobs2023deepspeed} and Ring-Attention~\citep{liu2023ring} incur \(\sim 40\text{–}120\,\mathrm{ms}\) cross-device latency—about \(20\text{–}40\times\) higher than our approach.

\paragraph{Performance-bound regime.}

To isolate algorithmic scaling, we evaluate the \emph{theoretical} latency of sequence parallelism and pipeline parallelism with communication removed (Fig.~\ref{fig: theor_cost}).
Our method, built on pipeline parallelism, attains a near-ideal acceleration by partitioning DiT blocks.

In contrast, sequence parallelism shows clear gains only at high resolutions; at moderate and low resolutions, it shifts the workload into a memory-bound regime, yielding little to no latency reduction and underutilizing compute.

\paragraph{Effectiveness of Stream Batch in the dual-pipeline scheduler.}

Figure~\ref{fig: streambatch} demonstrates that \emph{Stream Batch} substantially improves throughput, especially as the number of denoising steps increases.
More denoising steps create deeper in-flight pipelines, amplifying the benefit and delivering progressively larger throughput gains.

\section{Conclusions and New Outlooks}

We propose StreamDiffusionV2, which closes the gap between offline video diffusion and live streaming constrained in real-time with SLO constraints. Our training-free system couples an SLO-aware batching/block scheduler with a sink-token–guided rolling KV cache, a motion-aware noise controller, and a pipeline orchestration that parallelizes across denoising steps and network stages—delivering near-linear FPS scaling without violating latency. It runs on heterogeneous GPUs and flexible step counts, achieving 0.5 s TTFF and up to 58.28 FPS (14B) / 64.52 FPS (1.3B) on 4×H100, and maintaining high FPS even as steps increase. These results make state-of-the-art generative live streaming practical for both individual creators and enterprise platforms. 

\begin{figure}[t]
\centering
\begin{minipage}[t]{0.5\linewidth}
    \centering
    \includegraphics[width=\linewidth]{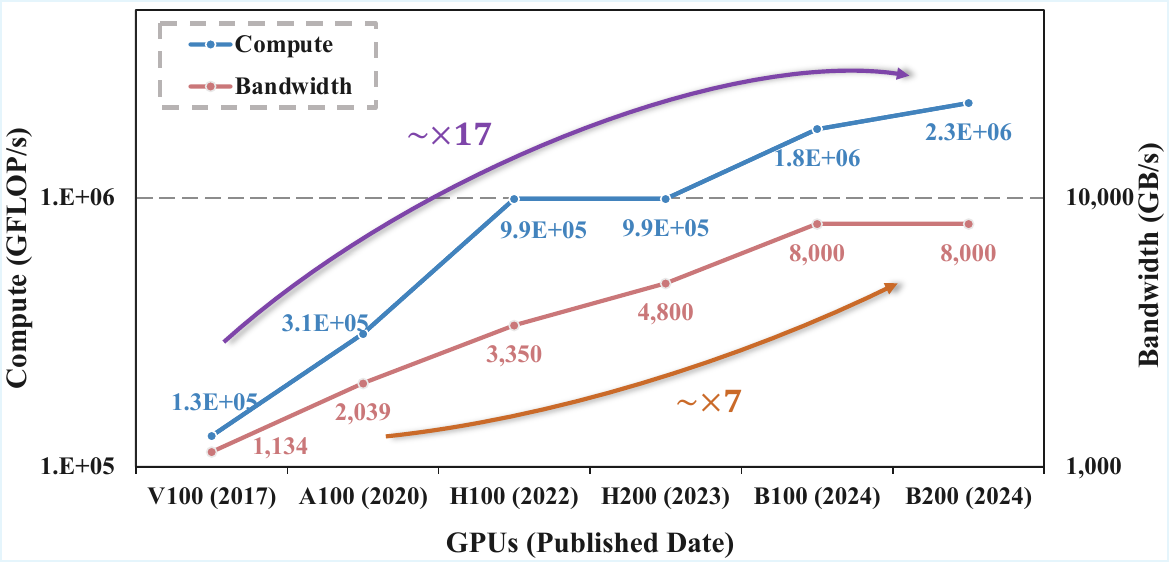}
    \vspace{-5pt}
    \caption{\textbf{The recent trends in the evolution of computational capability and memory bandwidth of AI hardware.}} 
    \label{fig: hardware}
\end{minipage}
\hfill
\begin{minipage}[t]{0.48\linewidth}
    \centering
    \includegraphics[width=\linewidth]{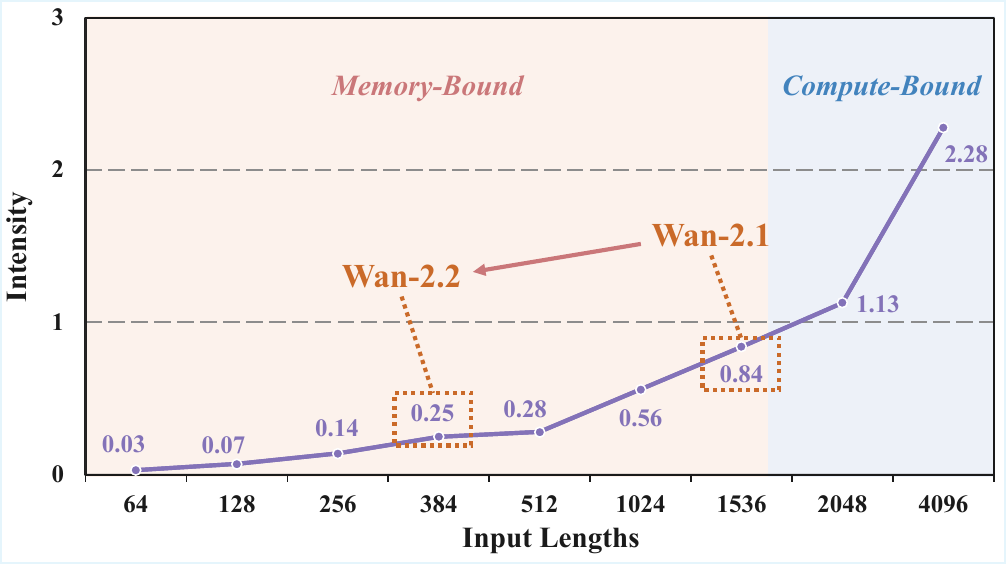}
    \vspace{-5pt}
    \caption{\textbf{Intensity of the auto-regressive video diffusion model when the input sequence length varies.}}
    \label{fig: algorithm_trend}
\end{minipage}
\vspace{-10pt}
\end{figure}

\noindent\textbf{Our new outlook.}
Video diffusion models have evolved from heavy bidirectional attention architectures toward autoregressive formulations, transforming a globally coupled spatiotemporal optimization into an amortized computation graph executed sequentially. From both hardware and algorithmic perspectives, this transition fundamentally reshapes the performance regime of video generation systems. We argue that future video models will increasingly operate in the \emph{memory-bound} region.

\begin{itemize}

    \item \textbf{Hardware trend.}
    Across recent GPU generations (Fig.~\ref{fig: hardware}), from V100 to A100, H100, and GB100-class accelerators, compute capability has scaled significantly faster than memory bandwidth. For example, moving from V100 (Tensor Core peak $\approx 130$ TFLOP/s) to GB100 (dense FP16/BF16 Tensor Core peak $\approx 2.3$ PFLOP/s), peak compute increases by $\sim 17\times$, whereas HBM bandwidth grows from $\sim 1.1$ TB/s to $\sim 8$ TB/s (only $\sim 7\times$). 

    Under the roofline model, this imbalance shifts the ``knee'' toward higher arithmetic intensity. Consequently, streaming inference workloads characterized by frequent memory movement and limited data reuse are increasingly constrained by memory bandwidth rather than raw compute. Therefore, the assumption that our serving pipeline is memory-sensitive is not transient; it becomes more pronounced as hardware continues to evolve.

    \item \textbf{Algorithm trend.}
    On the algorithmic side, modern video generation models are progressively adopting more compressed and structured latent representations (e.g., stronger VAEs and tokenization schemes) to enable longer temporal horizons and larger spatiotemporal contexts under fixed compute and memory budgets. 

    While compression reduces token counts and often improves reconstruction quality (e.g., Wan2.2 VAE reduces token counts to $\frac{1}{4}$ of Wan2.1), it also reduces arithmetic intensity by shrinking per-token compute relative to the required memory traffic in streaming inference. Our analysis shows that such compression can push the normalized arithmetic intensity to $\sim 0.2$ under the Wan2.2 VAE configuration (Fig.~\ref{fig: algorithm_trend}), further reinforcing the memory-bound regime in practical serving deployments.

\end{itemize}

Since both trends push the system deeper into the memory-bound regime, we believe that our approach is well-positioned to remain effective and potentially become more beneficial in future streaming systems by explicitly shaping memory traffic and scheduling under SLO constraints.

\clearpage
\newpage
\bibliographystyle{assets/plainnat}
\bibliography{example_paper}

@article{huang2025self,
  title={Self Forcing: Bridging the Train-Test Gap in Autoregressive Video Diffusion},
  author={Huang, Xun and Li, Zhengqi and He, Guande and Zhou, Mingyuan and Shechtman, Eli},
  journal={arXiv preprint arXiv:2506.08009},
  year={2025}
}

@inproceedings{yu2024repa,
    title={Representation Alignment for Generation: Training Diffusion Transformers Is Easier Than You Think},
    author={Sihyun Yu and Sangkyung Kwak and Huiwon Jang and Jongheon Jeong and Jonathan Huang and Jinwoo Shin and Saining Xie},
    year={2025},
    booktitle={International Conference on Learning Representations},
}

@inproceedings{yin2025slow,
  title={From slow bidirectional to fast autoregressive video diffusion models},
  author={Yin, Tianwei and Zhang, Qiang and Zhang, Richard and Freeman, William T and Durand, Fredo and Shechtman, Eli and Huang, Xun},
  booktitle={Proceedings of the Computer Vision and Pattern Recognition Conference},
  pages={22963--22974},
  year={2025}
}

@article{kodaira2023streamdiffusion,
  title={Streamdiffusion: A pipeline-level solution for real-time interactive generation},
  author={Kodaira, Akio and Xu, Chenfeng and Hazama, Toshiki and Yoshimoto, Takanori and Ohno, Kohei and Mitsuhori, Shogo and Sugano, Soichi and Cho, Hanying and Liu, Zhijian and Keutzer, Kurt},
  journal={arXiv preprint arXiv:2312.12491},
  year={2023}
}

@article{fang2024pipefusion,
  title={Pipefusion: Patch-level pipeline parallelism for diffusion transformers inference},
  author={Fang, Jiarui and Pan, Jinzhe and Wang, Jiannan and Li, Aoyu and Sun, Xibo},
  journal={arXiv preprint arXiv:2405.14430},
  year={2024}
}

@inproceedings{yeung2024balancing,
  title={Balancing Pipeline Parallelism with Vocabulary Parallelism},
  author={Tsung, Yeung Man and Qi, Penghui and Lin, Min and Wan, Xinyi},
  booktitle={Eighth Conference on Machine Learning and Systems}
}

@inproceedings{li2024distrifusion,
  title={Distrifusion: Distributed parallel inference for high-resolution diffusion models},
  author={Li, Muyang and Cai, Tianle and Cao, Jiaxin and Zhang, Qinsheng and Cai, Han and Bai, Junjie and Jia, Yangqing and Li, Kai and Han, Song},
  booktitle={Proceedings of the IEEE/CVF Conference on Computer Vision and Pattern Recognition},
  pages={7183--7193},
  year={2024}
}

@article{fang2024xdit,
  title={xDiT: an Inference Engine for Diffusion Transformers (DiTs) with Massive Parallelism},
  author={Fang, Jiarui and Pan, Jinzhe and Sun, Xibo and Li, Aoyu and Wang, Jiannan},
  journal={arXiv preprint arXiv:2411.01738},
  year={2024}
}

@article{jacobs2023deepspeed,
  title={Deepspeed ulysses: System optimizations for enabling training of extreme long sequence transformer models},
  author={Jacobs, Sam Ade and Tanaka, Masahiro and Zhang, Chengming and Zhang, Minjia and Song, Shuaiwen Leon and Rajbhandari, Samyam and He, Yuxiong},
  journal={arXiv preprint arXiv:2309.14509},
  year={2023}
}

@article{liu2023ring,
  title={Ring attention with blockwise transformers for near-infinite context},
  author={Liu, Hao and Zaharia, Matei and Abbeel, Pieter},
  journal={arXiv preprint arXiv:2310.01889},
  year={2023}
}

@article{yin2024improved,
  title={Improved distribution matching distillation for fast image synthesis},
  author={Yin, Tianwei and Gharbi, Micha{\"e}l and Park, Taesung and Zhang, Richard and Shechtman, Eli and Durand, Fredo and Freeman, Bill},
  journal={Advances in neural information processing systems},
  volume={37},
  pages={47455--47487},
  year={2024}
}

@article{salimans2022progressive,
  title={Progressive distillation for fast sampling of diffusion models},
  author={Salimans, Tim and Ho, Jonathan},
  journal={arXiv preprint arXiv:2202.00512},
  year={2022}
}

@article{kim2023consistency,
  title={Consistency trajectory models: Learning probability flow ode trajectory of diffusion},
  author={Kim, Dongjun and Lai, Chieh-Hsin and Liao, Wei-Hsiang and Murata, Naoki and Takida, Yuhta and Uesaka, Toshimitsu and He, Yutong and Mitsufuji, Yuki and Ermon, Stefano},
  journal={arXiv preprint arXiv:2310.02279},
  year={2023}
}

@article{lu2024simplifying,
  title={Simplifying, stabilizing and scaling continuous-time consistency models},
  author={Lu, Cheng and Song, Yang},
  journal={arXiv preprint arXiv:2410.11081},
  year={2024}
}

@inproceedings{meng2023distillation,
  title={On distillation of guided diffusion models},
  author={Meng, Chenlin and Rombach, Robin and Gao, Ruiqi and Kingma, Diederik and Ermon, Stefano and Ho, Jonathan and Salimans, Tim},
  booktitle={Proceedings of the IEEE/CVF conference on computer vision and pattern recognition},
  pages={14297--14306},
  year={2023}
}

@inproceedings{yin2024one,
  title={One-step diffusion with distribution matching distillation},
  author={Yin, Tianwei and Gharbi, Micha{\"e}l and Zhang, Richard and Shechtman, Eli and Durand, Fredo and Freeman, William T and Park, Taesung},
  booktitle={Proceedings of the IEEE/CVF conference on computer vision and pattern recognition},
  pages={6613--6623},
  year={2024}
}

@article{xie2024sana,
  title={Sana: Efficient high-resolution image synthesis with linear diffusion transformers},
  author={Xie, Enze and Chen, Junsong and Chen, Junyu and Cai, Han and Tang, Haotian and Lin, Yujun and Zhang, Zhekai and Li, Muyang and Zhu, Ligeng and Lu, Yao and others},
  journal={arXiv preprint arXiv:2410.10629},
  year={2024}
}

@article{chen2025sana,
  title={SANA-Video: Efficient Video Generation with Block Linear Diffusion Transformer},
  author={Chen, Junsong and Zhao, Yuyang and Yu, Jincheng and Chu, Ruihang and Chen, Junyu and Yang, Shuai and Wang, Xianbang and Pan, Yicheng and Zhou, Daquan and Ling, Huan and others},
  journal={arXiv preprint arXiv:2509.24695},
  year={2025}
}

@article{chen2024deep,
  title={Deep compression autoencoder for efficient high-resolution diffusion models},
  author={Chen, Junyu and Cai, Han and Chen, Junsong and Xie, Enze and Yang, Shang and Tang, Haotian and Li, Muyang and Lu, Yao and Han, Song},
  journal={arXiv preprint arXiv:2410.10733},
  year={2024}
}

@inproceedings{dalal2025one,
  title={One-minute video generation with test-time training},
  author={Dalal, Karan and Koceja, Daniel and Xu, Jiarui and Zhao, Yue and Han, Shihao and Cheung, Ka Chun and Kautz, Jan and Choi, Yejin and Sun, Yu and Wang, Xiaolong},
  booktitle={Proceedings of the Computer Vision and Pattern Recognition Conference},
  pages={17702--17711},
  year={2025}
}

@article{po2025long,
  title={Long-context state-space video world models},
  author={Po, Ryan and Nitzan, Yotam and Zhang, Richard and Chen, Berlin and Dao, Tri and Shechtman, Eli and Wetzstein, Gordon and Huang, Xun},
  journal={arXiv preprint arXiv:2505.20171},
  year={2025}
}

@article{gao2024matten,
  title={Matten: Video generation with mamba-attention},
  author={Gao, Yu and Huang, Jiancheng and Sun, Xiaopeng and Jie, Zequn and Zhong, Yujie and Ma, Lin},
  journal={arXiv preprint arXiv:2405.03025},
  year={2024}
}

@article{xi2025sparse,
  title={Sparse videogen: Accelerating video diffusion transformers with spatial-temporal sparsity},
  author={Xi, Haocheng and Yang, Shuo and Zhao, Yilong and Xu, Chenfeng and Li, Muyang and Li, Xiuyu and Lin, Yujun and Cai, Han and Zhang, Jintao and Li, Dacheng and others},
  journal={arXiv preprint arXiv:2502.01776},
  year={2025}
}

@article{yang2025sparse,
  title={Sparse VideoGen2: Accelerate Video Generation with Sparse Attention via Semantic-Aware Permutation},
  author={Yang, Shuo and Xi, Haocheng and Zhao, Yilong and Li, Muyang and Zhang, Jintao and Cai, Han and Lin, Yujun and Li, Xiuyu and Xu, Chenfeng and Peng, Kelly and others},
  journal={arXiv preprint arXiv:2505.18875},
  year={2025}
}

@inproceedings{zhang2025spargeattn,
  title={Spargeattn: Accurate sparse attention accelerating any model inference},
  author={Zhang, Jintao and Xiang, Chendong and Huang, Haofeng and Wei, Jia and Xi, Haocheng and Zhu, Jun and Chen, Jianfei},
  booktitle={International Conference on Machine Learning (ICML)},
  year={2025}
}

@inproceedings{zhang2025sageattention,
  title={SageAttention: Accurate 8-Bit Attention for Plug-and-play Inference Acceleration}, 
  author={Zhang, Jintao and Wei, Jia and Zhang, Pengle and Zhu, Jun and Chen, Jianfei},
  booktitle={International Conference on Learning Representations (ICLR)},
  year={2025}
}

@inproceedings{zhang2024sageattention2,
  title={Sageattention2: Efficient attention with thorough outlier smoothing and per-thread int4 quantization},
  author={Zhang, Jintao and Huang, Haofeng and Zhang, Pengle and Wei, Jia and Zhu, Jun and Chen, Jianfei},
  booktitle={International Conference on Machine Learning (ICML)},
  year={2025}
}

@article{chen2024delta,
  title={$\Delta$-DiT: A Training-Free Acceleration Method Tailored for Diffusion Transformers},
  author={Chen, Pengtao and Shen, Mingzhu and Ye, Peng and Cao, Jianjian and Tu, Chongjun and Bouganis, Christos-Savvas and Zhao, Yiren and Chen, Tao},
  journal={arXiv preprint arXiv:2406.01125},
  year={2024}
}

@article{selvaraju2024fora,
  title={Fora: Fast-forward caching in diffusion transformer acceleration},
  author={Selvaraju, Pratheba and Ding, Tianyu and Chen, Tianyi and Zharkov, Ilya and Liang, Luming},
  journal={arXiv preprint arXiv:2407.01425},
  year={2024}
}

@inproceedings{wimbauer2024cache,
  title={Cache me if you can: Accelerating diffusion models through block caching},
  author={Wimbauer, Felix and Wu, Bichen and Schoenfeld, Edgar and Dai, Xiaoliang and Hou, Ji and He, Zijian and Sanakoyeu, Artsiom and Zhang, Peizhao and Tsai, Sam and Kohler, Jonas and others},
  booktitle={Proceedings of the IEEE/CVF Conference on Computer Vision and Pattern Recognition},
  pages={6211--6220},
  year={2024}
}

@article{liu2024timestep,
  title={Timestep Embedding Tells: It's Time to Cache for Video Diffusion Model},
  author={Liu, Feng and Zhang, Shiwei and Wang, Xiaofeng and Wei, Yujie and Qiu, Haonan and Zhao, Yuzhong and Zhang, Yingya and Ye, Qixiang and Wan, Fang},
  journal={arXiv preprint arXiv:2411.19108},
  year={2024}
}

@article{zhou2025less,
  title={Less is Enough: Training-Free Video Diffusion Acceleration via Runtime-Adaptive Caching},
  author={Zhou, Xin and Liang, Dingkang and Chen, Kaijin and Feng, Tianrui and Chen, Xiwu and Lin, Hongkai and Ding, Yikang and Tan, Feiyang and Zhao, Hengshuang and Bai, Xiang},
  journal={arXiv preprint arXiv:2507.02860},
  year={2025}
}

@article{shih2023parallel,
  title={Parallel sampling of diffusion models},
  author={Shih, Andy and Belkhale, Suneel and Ermon, Stefano and Sadigh, Dorsa and Anari, Nima},
  journal={Advances in Neural Information Processing Systems},
  volume={36},
  pages={4263--4276},
  year={2023}
}

@article{yang2025longlive,
  title={LongLive: Real-time Interactive Long Video Generation},
  author={Yang, Shuai and Huang, Wei and Chu, Ruihang and Xiao, Yicheng and Zhao, Yuyang and Wang, Xianbang and Li, Muyang and Xie, Enze and Chen, Yingcong and Lu, Yao and others},
  journal={arXiv preprint arXiv:2509.22622},
  year={2025}
}

@article{chen2025skyreels,
  title={Skyreels-v2: Infinite-length film generative model},
  author={Chen, Guibin and Lin, Dixuan and Yang, Jiangping and Lin, Chunze and Zhu, Junchen and Fan, Mingyuan and Zhang, Hao and Chen, Sheng and Chen, Zheng and Ma, Chengcheng and others},
  journal={arXiv preprint arXiv:2504.13074},
  year={2025}
}

@article{kodaira2025streamdit,
  title={Streamdit: Real-time streaming text-to-video generation},
  author={Kodaira, Akio and Hou, Tingbo and Hou, Ji and Tomizuka, Masayoshi and Zhao, Yue},
  journal={arXiv preprint arXiv:2507.03745},
  year={2025}
}

@article{teng2025magi,
  title={MAGI-1: Autoregressive Video Generation at Scale},
  author={Teng, Hansi and Jia, Hongyu and Sun, Lei and Li, Lingzhi and Li, Maolin and Tang, Mingqiu and Han, Shuai and Zhang, Tianning and Zhang, WQ and Luo, Weifeng and others},
  journal={arXiv preprint arXiv:2505.13211},
  year={2025}
}

@article{wan2025wan,
  title={Wan: Open and advanced large-scale video generative models},
  author={Wan, Team and Wang, Ang and Ai, Baole and Wen, Bin and Mao, Chaojie and Xie, Chen-Wei and Chen, Di and Yu, Feiwu and Zhao, Haiming and Yang, Jianxiao and others},
  journal={arXiv preprint arXiv:2503.20314},
  year={2025}
}

@article{he2025matrix,
    title={Matrix-Game 2.0: An Open-Source, Real-Time, and Streaming Interactive World Model},
    author={He, Xianglong and Peng, Chunli and Liu, Zexiang and Wang, Boyang and Zhang, Yifan and Cui, Qi and Kang, Fei and Jiang, Biao and An, Mengyin and Ren, Yangyang and Xu, Baixin and Guo, Hao-Xiang and Gong, Kaixiong and Wu, Cyrus and Li, Wei and Song, Xuchen and Liu, Yang and Li, Eric and Zhou, Yahui},
    journal={arXiv preprint arXiv:2508.13009},
    year={2025}
  }

@inproceedings{valevskidiffusion,
  title={Diffusion Models Are Real-Time Game Engines},
  author={Valevski, Dani and Leviathan, Yaniv and Arar, Moab and Fruchter, Shlomi},
  booktitle={The Thirteenth International Conference on Learning Representations}
}

@article{liang2024looking,
  title={Looking Backward: Streaming Video-to-Video Translation with Feature Banks},
  author={Liang, Feng and Kodaira, Akio and Xu, Chenfeng and Tomizuka, Masayoshi and Keutzer, Kurt and Marculescu, Diana},
  journal={arXiv preprint arXiv:2405.15757},
  year={2024}
}

@article{sripanidkulchai2004feasibility,
  title={The feasibility of supporting large-scale live streaming applications with dynamic application end-points},
  author={Sripanidkulchai, Kunwadee and Ganjam, Aditya and Maggs, Bruce and Zhang, Hui},
  journal={ACM SIGCOMM computer communication review},
  volume={34},
  number={4},
  pages={107--120},
  year={2004},
  publisher={ACM New York, NY, USA}
}

@inproceedings{zhang2005coolstreaming,
  title={CoolStreaming/DONet: A data-driven overlay network for peer-to-peer live media streaming},
  author={Zhang, Xinyan and Liu, Jiangchuan and Li, Bo and Yum, Y-SP},
  booktitle={Proceedings IEEE 24th Annual Joint Conference of the IEEE Computer and Communications Societies.},
  volume={3},
  pages={2102--2111},
  year={2005},
  organization={IEEE}
}

@article{huang2008challenges,
  title={Challenges, design and analysis of a large-scale p2p-vod system},
  author={Huang, Yan and Fu, Tom ZJ and Chiu, Dah-Ming and Lui, John CS and Huang, Cheng},
  journal={ACM SIGCOMM computer communication review},
  volume={38},
  number={4},
  pages={375--388},
  year={2008},
  publisher={ACM New York, NY, USA}
}

@inproceedings{bain2021frozen,
  title={Frozen in time: A joint video and image encoder for end-to-end retrieval},
  author={Bain, Max and Nagrani, Arsha and Varol, G{\"u}l and Zisserman, Andrew},
  booktitle={Proceedings of the IEEE/CVF international conference on computer vision},
  pages={1728--1738},
  year={2021}
}

@inproceedings{chen2024panda,
  title={Panda-70m: Captioning 70m videos with multiple cross-modality teachers},
  author={Chen, Tsai-Shien and Siarohin, Aliaksandr and Menapace, Willi and Deyneka, Ekaterina and Chao, Hsiang-wei and Jeon, Byung Eun and Fang, Yuwei and Lee, Hsin-Ying and Ren, Jian and Yang, Ming-Hsuan and others},
  booktitle={Proceedings of the IEEE/CVF Conference on Computer Vision and Pattern Recognition},
  pages={13320--13331},
  year={2024}
}

@article{yang2024vript,
  title={Vript: A video is worth thousands of words},
  author={Yang, Dongjie and Huang, Suyuan and Lu, Chengqiang and Han, Xiaodong and Zhang, Haoxin and Gao, Yan and Hu, Yao and Zhao, Hai},
  journal={Advances in Neural Information Processing Systems},
  volume={37},
  pages={57240--57261},
  year={2024}
}

@inproceedings{teed2020raft,
  title={Raft: Recurrent all-pairs field transforms for optical flow},
  author={Teed, Zachary and Deng, Jia},
  booktitle={European conference on computer vision},
  pages={402--419},
  year={2020},
  organization={Springer}
}

@inproceedings{radford2021learning,
  title={Learning transferable visual models from natural language supervision},
  author={Radford, Alec and Kim, Jong Wook and Hallacy, Chris and Ramesh, Aditya and Goh, Gabriel and Agarwal, Sandhini and Sastry, Girish and Askell, Amanda and Mishkin, Pamela and Clark, Jack and others},
  booktitle={International conference on machine learning},
  pages={8748--8763},
  year={2021},
  organization={PmLR}
}

@article{oquab2024dinov2,
  title={DINOv2: Learning Robust Visual Features without Supervision},
  author={Oquab, Maxime and Darcet, Timoth{\'e}e and Moutakanni, Th{\'e}o and Vo, Huy and Szafraniec, Marc and Khalidov, Vasil and Fernandez, Pierre and Haziza, Daniel and Massa, Francisco and El-Nouby, Alaaeldin and others},
  journal={Transactions on Machine Learning Research Journal},
  pages={1--31},
  year={2024}
}

@article{dao2022flashattention,
  title={Flashattention: Fast and memory-efficient exact attention with io-awareness},
  author={Dao, Tri and Fu, Dan and Ermon, Stefano and Rudra, Atri and R{\'e}, Christopher},
  journal={Advances in neural information processing systems},
  volume={35},
  pages={16344--16359},
  year={2022}
}

\clearpage
\newpage
\beginappendix

\section{NVIDIA H100 SXM Roofline Parameters}
We derive the roofline parameters for the NVIDIA H100 SXM GPU. The dense FP16 throughput is $990~\mathrm{TFLOP/s}$ and GPU memory bandwidth is $1.5~\mathrm{TB/s}$, which are derived from empirical measurements and hardware specifications~\footnote{https://resources.nvidia.com/en-us-gpu-resources/h100-datasheet-24306}. The ridge arithmetic intensity is 
\begin{equation}
\mathrm{AI}_\mathrm{ridge}=\frac{990~\mathrm{TFLOP/s}}{1.5~\mathrm{TB/s}}=660~\mathrm{FLOP/Byte}
\end{equation}


\section{Hardware Intensity Analysis and Configuration}
\label{appendix_B}
\begin{wrapfigure}{r}{0.5\textwidth}
\vspace{-0.5cm}
\begin{center}
\includegraphics[width=0.48\textwidth]{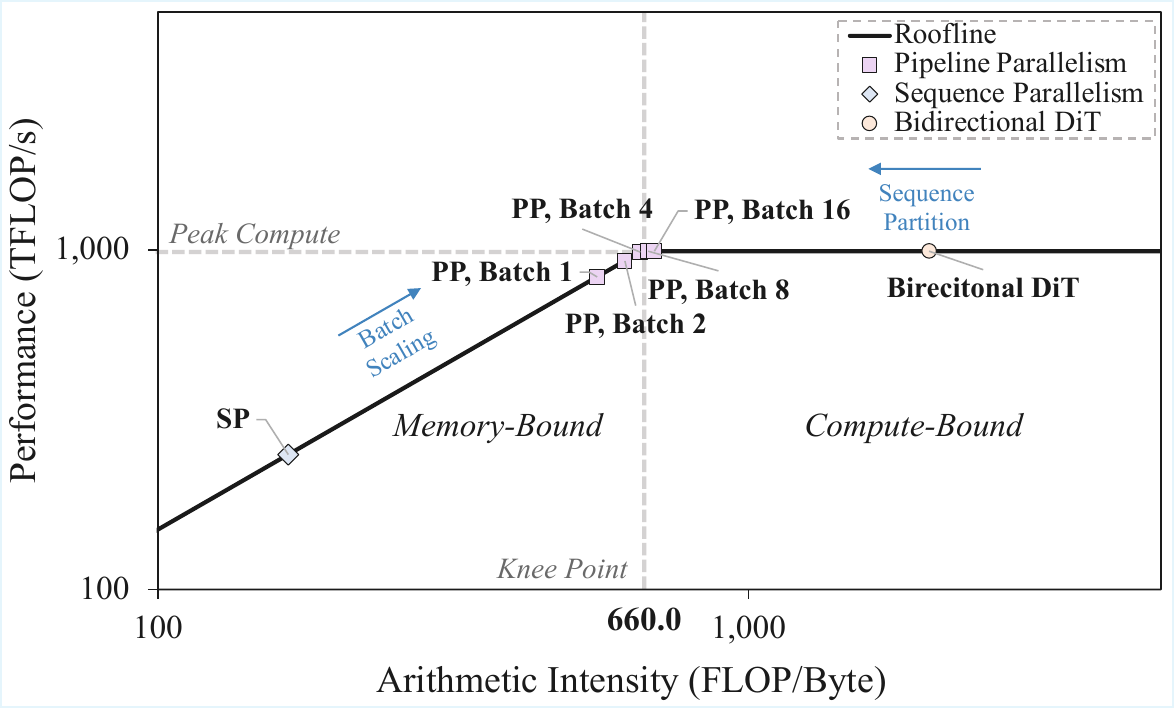}
\end{center}
\caption{\textbf{Roofline analysis of sequence parallelism and our pipeline orchestration.} We compare the Sequence Parallelism and Pipeline Parallelism under varying batch sizes in the causal DiT, compared with the bidirectional DiT. The results demonstrate that our approach operates near the knee point of the roofline, effectively avoiding compute under-utilization as seen in the bidirectional DiT and memory bandwidth limitations in Sequence Parallelism. The model is profiled on an NVIDIA H100 SXM GPU with a peak performance of 990 TFLOP/s and a knee point at an arithmetic intensity (AI) of 660.0 FLOP/Byte. The token length is 1,560 (a 4-frame chunk at 480P resolution).}
\vspace{-0.4cm}
\label{fig: roofline}
\end{wrapfigure}
Consider a DiT backbone with hidden dimension $C$, query length $L_q$, KV-cache length $L_{kv}$ (assuming full KV read/write each step), batch size $b$, and $N$ layers. The total compute cost $F$ is defined as:
\begin{equation}
    F = N\left(24b L_q C^2 + 4b L_q L_{kv} C\right)
\end{equation}
For memory movement $B$, we account for: (i) activations, (ii) model weights, and (iii) KV-cache traffic. Furthermore, we include additional system-level memory operations, such as sliding-window buffering and feature concatenation. This yields:
\begin{equation}
    B = N\left(24C^2 + 4b L_{kv}C + 4b L_qC + 36b L_qC\right) + b \gamma
\end{equation}
where $\gamma$ captures the extra memory movement beyond the standard DiT read/write pattern. The normalized hardware intensity is then defined as:

Using a practical configuration on an NVIDIA H100 with 480p video input, we set $C=2,048$, $L_q=1,536$, $L_{kv}=1,536 \times 6$, and $N=30$. Importantly, using NVIDIA Nsight Systems, we directly measure the total memory traffic during DiT inference---corresponding to $\gamma$---to be 5.2~GiB, while the realized device bandwidth is approximately 1.5~TB/s. 

The practical intensity is calculated as follows (using FP16 dense peak $\approx$ 990~TFLOPS as an upper bound, noting that the effective compute peak is lower in practice):
\begin{equation}
    \text{Intensity} \approx \frac{F / 990\,\text{TFLOPS}}{B / 1.5\,\text{TB/s}} \approx 0.84 < 1
\end{equation}

This result indicates a \textbf{memory-bound regime}. This conclusion is consistent with our empirical observations in Fig. 15 (batch denoising), where performance scales with memory bandwidth rather than peak computation.

\section{REPA finetuning for quality enhancement}
StreamDiffusionV2 is a training-free system solution to transform efficient video models for live streaming applications. Orthogonal to the training-free solution, we employ the REPA~\citep{yu2024repa} training strategy during Causal-DiT distillation~\citep{yin2025slow} to enhance the video quality. The alignment loss is denoted by
\begin{equation*}
\mathcal{L}_{\mathrm{REPA}}(\theta, \phi) := 
- \mathbb{E}_{\mathbf{x}_*, \epsilon, t} \left[ 
    \frac{1}{N} \sum_{n=1}^N 
    \mathrm{sim}\!\left( \mathbf{y}_*^{[n]}, \, h_{\phi}\!\left(\mathbf{h}_t^{[n]}\right) \right) 
\right],
\end{equation*}
where $\mathbf{y}_*^{[n]}$ indicates the output of DINOv2~\citep{oquab2024dinov2}, $\mathbf{h}_t^{[n]}$ represents the hidden state of DiT in timestep $t$, $h_{\phi}$ is the projection network parameterized by $\phi$, and $\mathrm{sim}(\cdot,\cdot)$ denotes cosine similarity. When combined with the original DMD~\citep{yin2024one} distillation objective $\mathcal{L}_{\mathrm{DMD}}$, the overall training objective is defined as
\begin{equation*}
\mathcal{L} = \mathcal{L}_{\mathrm{DMD}} + \lambda \mathcal{L}_{\mathrm{REPA}},
\end{equation*}
where $\lambda$ is a scaling factor.

\begin{figure}[t]
\centering
\begin{minipage}[h]{0.48\linewidth}
    \centering
    \includegraphics[width=\linewidth]{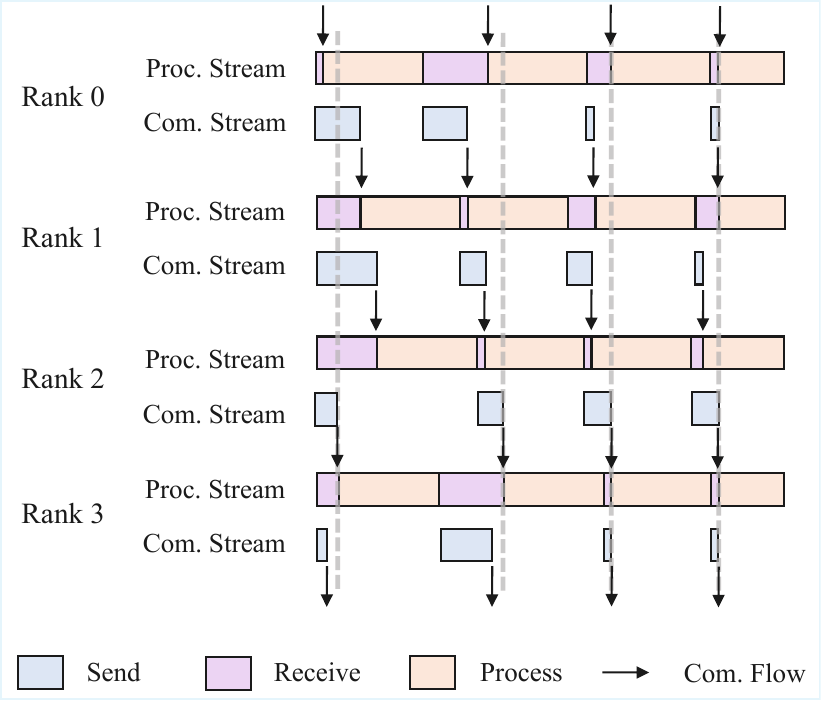}
    \vspace{-15pt}
    \caption{\textbf{Execution timeline of the Pipeline-orchestration architecture.}}
    \label{fig: timeline}
    \vspace{-10pt}
\end{minipage}
\hfill
\begin{minipage}[t]{0.48\linewidth}
    \centering
    \includegraphics[width=\linewidth]{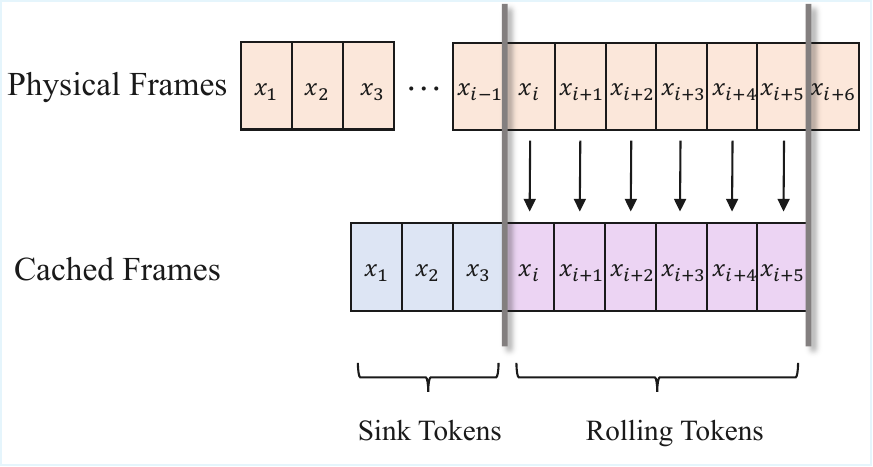}
    \vspace{-15pt}
    \caption{\textbf{The detailed illustration of the Rolling KV Cache and Sink Token designs.}}
    \label{fig: kv_cache}
    \vspace{-10pt}
\end{minipage}
\vspace{-10pt}
\end{figure}

\begin{figure}[t]
\centering
\includegraphics[width=0.75\linewidth]{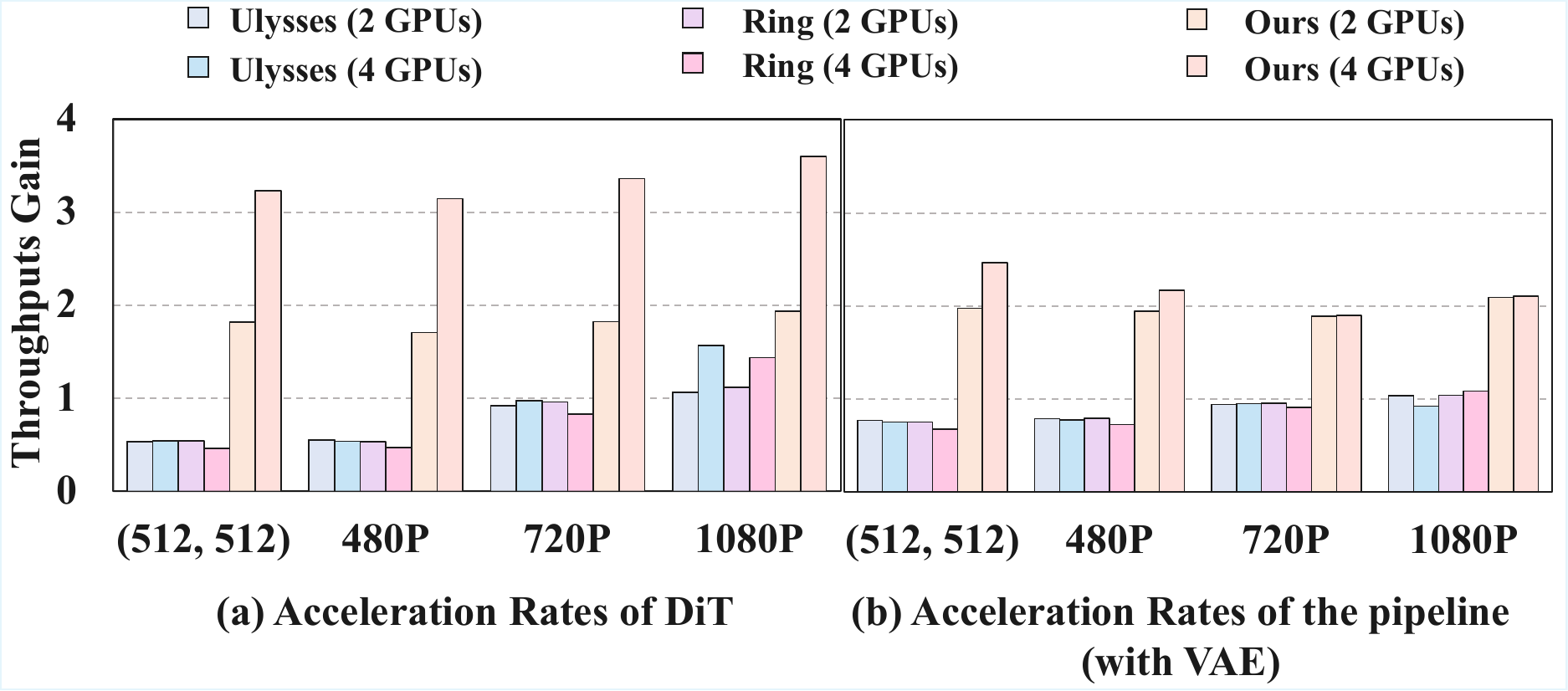}
\vspace{-10pt}
\caption{\textbf{Acceleration rate of different approaches on various resolutions.} \textit{Left}: Testing the acceleration rate of DiT only. \textit{Right}: Testing the acceleration of the whole pipeline (with VAE).}
\vspace{-10pt}
\label{fig: sp_inference}
\end{figure}

\section{Figure illustration}
\label{appendix_D}
\textbf{Figure overview.}
\textbf{Fig.~\ref{fig: timeline}} shows how we minimize pipeline bubbles by separating compute and communication into two concurrent streams, overlapping kernels with P2P transfers.

\textbf{Fig.~\ref{fig: kv_cache}} illustrates the rolling KV cache with sink tokens for consistent, frame-by-frame updates.

\textbf{Fig.~\ref{fig: sp_inference}} reports a comparison of parallelization strategies (e.g., sequence parallelism vs.\ our pipeline scheduler).

\end{document}